# Title

**AI-guided stimuli discovery and generation to optimize facial emotion perception studies in autism**

**Authors**

Kushin Mukherjee[1,2]*, Na Yeon Kim[3]*, Maren Wehrheim[1,4], Ralph Adolphs[5], and Kohitij Kar [1,4]

**Affiliation**

1. Centre for Integrative and Applied Neuroscience, York University, Canada
2. Department of Psychology, Stanford University, USA
3. Department of Psychology, University of California, Riverside, USA
4. Department of Biology, and Centre for Vision Research, Toronto, York University, Canada
5. Division of the Humanities and Social Sciences, California Institute of Technology, Pasadena, CA, USA

* contributed equally to this work.

Correspondence should be addressed to Kohitij Kar
E-mail: k0h1t1j@yorku.ca

**Conflict of interests**

The author declares no competing financial interests.

**Acknowledgments**

KK has been supported by funds from the Canada Research Chair Program, the Simons Foundation Autism Research Initiative (SFARI, 967073), Brain-Canada Foundation (2023-0259), the Canada First Research Excellence Funds (VISTA Program), the National Sciences and Engineering Research Council of Canada (NSERC, RGPIN-2024-06223), and the Ontario Early Researcher Awards program (ER24-18-272). NYK was supported by the Della Martin Foundation Postdoctoral Fellowship. RA was supported by the Simons Foundation Autism Research Initiative (SFARI, 990500).

## Data and Code Availability

Curated de-identified data and code for reproducing the reported figure panels will be available at https://github.com/vital-kolab/autism-ann-emotion upon publication.

# Abstract

Understanding how perceptual differences arise between autistic and neurotypical individuals remains a major challenge in autism research. A key obstacle is the lack of behavioral assays that are both sensitive to these differences and sufficiently mechanistic to reveal the computations that underlie them. Facial emotion perception is a compelling domain for making progress on assay development because differences between autistic and neurotypical observers have been widely reported, yet the mechanisms that produce these differences remain unresolved. Some studies find robust group differences in emotion judgments, whereas others report weak or inconsistent effects, suggesting that the behavioral phenotype is not expressed uniformly across tasks, individuals, or stimuli. Here, we show that autistic–neurotypical differences in facial emotion perception are not uniformly distributed across images, but are concentrated in a small subset of diagnostic stimuli. We trained population-specific artificial neural network (ANN) models to predict image-level emotion judgments in autistic and neurotypical participants and used these models to prospectively screen novel facial expressions. ANN-guided image selection identified stimuli that produced larger autistic–neurotypical behavioral separation than randomly selected images in an independent cohort, demonstrating that computational models can improve assay sensitivity by selecting high-leverage stimuli. We then extended this framework from stimulus selection to stimulus transformation. Using a generative adversarial network constrained by the ANN-based behavioral models, we synthesized modified versions of diagnostic images designed to move responses toward greater autistic–neurotypical agreement. In a phenotype-matched validation analysis, these synthesized images reduced behavioral separation relative to their matched base images. Together, these findings establish a model-guided framework for discovering and transforming stimuli that reveal population-specific perceptual differences. More broadly, this approach shifts behavioral phenotyping from averaging across fixed stimulus sets toward computationally optimized assays that identify the stimulus conditions under which neurodivergent perception diverges or converges.

# Introduction

Autism is clinically defined by differences in social communication [1] and restricted or repetitive patterns of behavior, but the perceptual and cognitive computations that underlie these differences[2–6] remain difficult to measure precisely. This challenge is especially clear in the study of facial emotion perception. Facial expressions provide a controlled and behaviorally tractable way to study how observers extract socially relevant information from visual input, and many studies have examined how autistic and neurotypical individuals recognize, discriminate, or rate facial expressions [7,8]. Across this literature, group differences have often been reported in overall accuracy [9], sensitivity[10], reaction time [11], gaze allocation [12,13], or response patterns[7]. Converging neural evidence also supports the possibility that facial information is processed differently in autism[14,15], including neuroimaging studies of face-selective cortex[16] and single-neuron recordings[17,18] showing atypical responses to faces in autistic individuals. However, the empirical picture remains heterogeneous. Meta-analyses have shown that the magnitude and reliability of autistic–neurotypical differences depend strongly on the task, stimulus set, emotion category, response format, and participant characteristics [19]. Some studies report robust differences, whereas others find small or absent effects. This variability has limited the use of facial emotion tasks as stable behavioral assays and has made it difficult to determine whether inconsistent findings reflect true heterogeneity in autistic perceptual phenotypes, limitations of experimental design, or both. An important source of heterogeneity is the stimulus set itself: many facial-emotion studies rely on constrained, historically dominant image sets, such as Ekman-style posed expressions, which may differ in identity sampling, image statistics, expression intensity, and ecological validity. Thus, inconsistent findings may reflect not only participant heterogeneity, but also stimulus-set bias and uneven diagnostic sensitivity across individual images.

This issue is compounded by the fact that most behavioral assays summarize performance after averaging over many images. In a typical experiment, facial stimuli are selected to sample a predefined experimental dimension, such as emotion category, identity, intensity, or morph level. Responses are then averaged across images to estimate group-level accuracy, psychometric slope, category boundary, or mean rating. This approach is appropriate when the goal is to estimate an overall effect across a stimulus class, but it implicitly treats individual images as exchangeable samples from that class. This assumption may be problematic for autism research. If autistic–neurotypical differences are expressed only for particular images because of specific combinations of identity, expression intensity, local facial features, or higher-order visual structure, then averaging across the full stimulus set will dilute the very effects the experiment aims to measure. Under this view, weak or inconsistent group effects do not necessarily imply that facial emotion perception is unaffected in autism. They may instead indicate that diagnostic information is sparse at the level of individual images and is therefore lost when stimuli are pooled.

Heterogeneity arises both at the level of the participants as well as at the level of stimuli. Although constrained by sample sizes, several studies have explored individual differences across participants, for instance by parameterizing autism as a continuous dimension (e.g., from autistic trait scores) [20,21]. Here we focus on the second problematic source of heterogeneity: the individuality of the stimuli themselves. Rather than asking only whether autistic and neurotypical observers differ on average across a stimulus set, we ask which specific images elicit the largest behavioral divergence, whether those images share a computable structure, and whether that structure can be predicted. This formulation treats stimulus variability not as noise to be averaged away, but as a source of mechanistic information. A facial image that produces similar responses in autistic and neurotypical observers is not equivalent, for the purposes of phenotyping, to one that produces large group divergence. Identifying the latter class of images requires models that operate at the level of individual stimuli[22] and generate quantitative predictions before new behavioral data are collected. If successful, this approach would not only provide us with a better mechanistic understanding of which facial features drive the observed group differences, but would also permit us to synthesize new stimuli that could better distinguish groups and be used in future stimulus sets.

Image-computable models[23,22] provide a principled route to this goal. Such models take the image itself as input, extract a feature representation, and predict behavioral responses on an image-by-image basis. Artificial neural networks (ANNs) are particularly useful in this context because they provide high-dimensional visual representations that have been linked to human visual recognition behavior[24], neural responses in the primate ventral visual pathway[25,26], including the macaque face-processing system[27], and facial expression judgments[28,29]. Importantly, ANNs need not serve merely as generic predictors of human behavior. When combined with population-specific readouts, the same visual representation can be used to model how different populations transform image information into behavioral judgments[30]. This makes it possible to ask whether autistic and neurotypical observers differ not only in overall performance, but in the visual features that drive their decisions.

Prior work[30,31] suggests that this strategy may be especially informative for facial emotion perception. ANN-based models of facial affect can capture image-level structure in neurotypical emotion judgments[29], and previous comparisons indicate that such models align more closely with neurotypical than autistic behavioral patterns[30]. This asymmetry is not simply a limitation of the model. If an ANN captures visual features that predict neurotypical judgments but fails to predict autistic judgments for particular images, those failures may reveal where autistic and neurotypical observers rely on different image information. Conversely, if separate population-specific decoders trained on the same ANN representation predict different responses to the same image, the divergence between those predictions can be used to identify stimuli that are likely to be behaviorally diagnostic. This provides a route from retrospective model fitting to prospective experimental design.

Here, we test whether behavior-aligned ANNs can prospectively discover facial expressions that can maximally distinguish autistic–neurotypical behavior. We first reanalyze an existing facial emotion dataset[10] to determine whether group differences are broadly distributed across images or concentrated in a sparse subset of stimuli. We then train population-specific ANN models to predict image-level emotion judgments separately for autistic and neurotypical participants. Applying these models to a new facial-expression dataset, we ask whether they can identify, before behavioral testing, the images most likely to produce large group differences in an independent cohort. This prospective test is critical: a model that merely explains past data may be unhelpful for experimental design, whereas a model that selects diagnostic images for new participants can serve as an active tool for assay construction.

We then extend the framework from stimulus selection[32] to stimulus synthesis[26,33]. Selection alone tests whether a model can find diagnostic images within an existing stimulus set. Synthesis, however, provides a stronger test by asking whether the model can guide targeted image transformations that alter the behavioral gap itself. To do this, we integrate the population-specific behavioral models with a generative adversarial network[34] that supports controlled manipulation of facial expressions while preserving realistic facial structure and identity. Starting from diagnostic base images, the closed-loop procedure searches the generative image space for modified expressions predicted to reduce autistic–neurotypical divergence. We then test whether these model-guided transformations reduce the behavioral gap in an independent validation cohort. This experiment asks whether the model captures behaviorally relevant image structure well enough to intervene on the stimulus and change the measured group difference.

Taken together, this study develops an image-computable framework for precision behavioral assay design in autism research. The goal is not to classify individuals, define a single autistic phenotype, or imply that facial emotion perception alone explains differences in social communication. Instead, using facial emotion judgments as a controlled test case for a broader methodological problem, we provide a template to demonstrate: how to identify the stimuli that reveal reliable and interpretable behavioral variation across neurodevelopmental populations. By combining population-specific behavioral modeling, prospective image selection, and closed-loop image synthesis, we show how ANN-guided experiments can transform stimulus variability from a source of inconsistency into a tool for discovery. More broadly, this approach suggests that brain- and behavior-aligned computational models can move beyond retrospective explanation and become active instruments for experimental design [35] in clinical and translational neuroscience.

# Results

As discussed above, prior approaches[7,9,10] have primarily relied on stimulus-averaged behavioral measures. To evaluate whether group differences instead emerge from image-specific effects, we conducted an image-level reanalysis of existing data from Wang and Adolphs (2017)[10]. We chose this dataset as the starting point for three reasons. First, its original psychometric results were especially compelling for testing image-level autistic–neurotypical differences. Second, the participant groups were well characterized and closely matched relative to many available behavioral datasets. Third, the stimuli were tightly controlled: images contained only face information, excluded hair and background, and were equated for luminance and spectral energy. This made the dataset suitable as a controlled springboard for asking whether diagnostic information is sparse at the image level.

### Behavioral differences are often driven by a small subset of diagnostic images

Motivated by the possibility that averaging across stimuli obscures meaningful variability, we reanalyzed the dataset from *Wang and Adolphs (2017)*[10], in which autistic (n = 18) and neurotypical (NT) adults (n = 15) judged whether briefly presented (1000 ms) faces appeared *happy* or *fearful* (**Figure 1A**). For each image, we computed the cross-validated difference in mean "happy" judgments between the two groups, selecting diagnostic images in one subset of 7 participants and validating them on the held-out subjects to avoid circularity. As shown in **Figure 1B**, the number of discriminative images decreased sharply as the group difference (Δ behavior = autistic – NT) increased. Only a small number of images produced substantial NT–ASD behavioral separations, indicating that overall group effects were driven by a small subset of "high-leverage" stimuli rather than by uniform perceptual shifts across all images. This finding motivates our first goal: to develop an image-computable framework capable of *predicting, a priori*, which new images are likely to evoke large NT–ASD behavioral gaps (**Figure 1C**). Artificial neural networks (ANNs) trained on existing emotion-judgment data can serve as such predictive engines, enabling closed-loop experimental design in which future stimuli are selected to maximize sensitivity to neurodivergent processing.

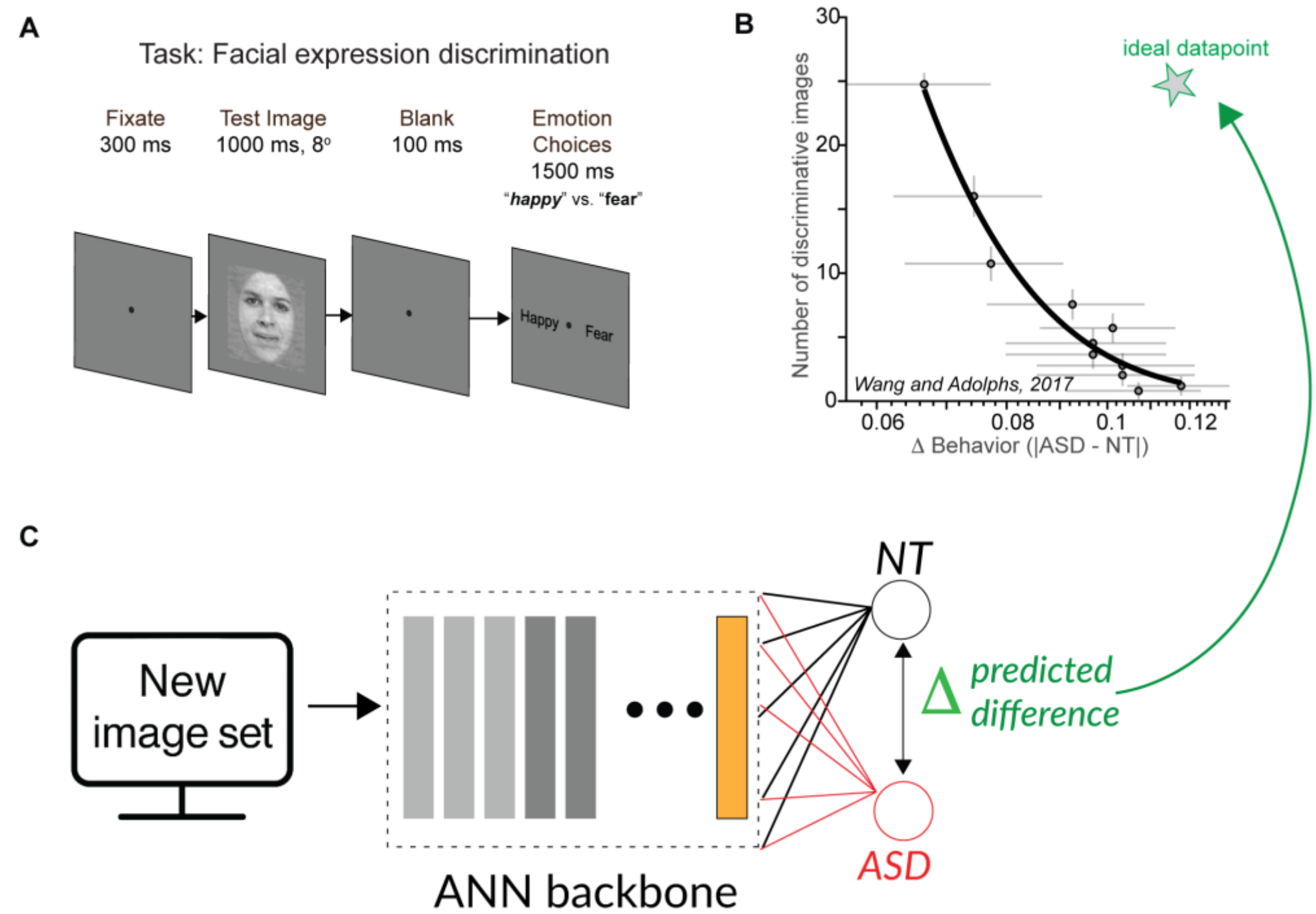

**Figure 1. Differences in facial emotion discrimination emerge from a small subset of diagnostic images. A.** Task schematic from *Wang & Adolphs (2017)*. Participants fixated (300 ms), viewed a test face (1000 ms), followed by a blank (100 ms) and two emotion choices ("happy" vs. "fear"; 1500 ms). **B.** Cross-validated distribution of image-wise behavioral differences (Δ behavior = autistic – NT). The number of discriminative images declines exponentially as the between-group difference increases. **C.** Conceptual goal of **ANN-driven closed-loop stimulus selection**: models trained on prior data predict the behavioral difference (Δ prediction) for novel images, guiding future experiments toward the *ideal datapoint* that maximizes autistic–NT separation.

## ANN models trained on behavioral data predict diagnostic images in new participants

We next asked whether image-computable models trained on existing behavioral data could predict which novel facial expressions would produce large autistic–neurotypical differences. To do this, we trained a suite of population-specific ANN models using the Wang and Adolphs dataset[10] (**Figure 2A**). This analysis was possible because image-level behavior in the Wang and Adolphs dataset, while correlated across autistic and neurotypical participants, was not at the noise ceiling (**Figure S1**), indicating that the two groups shared a broad emotion-judgment structure while still differing systematically for particular images. Each ANN backbone—spanning convolutional (AlexNet[36], VGG-19[37], ResNet-50[38], ConvNeXt[39]), recurrent (CORNet-S[40]), transformer (ViT[41]), and contrastive (CLIP[42]) architectures—was used as a fixed visual encoder. We then evaluated these models on an *independent* facial expression dataset, the *Montreal Set of Facial Displays of Emotion* (MSFDE[43]), comprising eight identities across five emotional intensities (**Figure 2B**). This provided a strong generalization test because MSFDE images differed from the Wang and Adolphs training stimuli in several respects: they depicted different identities, included hair and more natural face-boundary information, varied in shading and resolution, and were drawn from a different stimulus source while still parametrically sampling facial emotion. Before fixing a single feature layer for stimulus selection, we used this MSFDE generalization test to ask which stages of the visual hierarchy carried behaviorally useful diagnostic information. For each model, we extracted feature activations from representative internal layers spanning early, middle, and late processing stages and passed these activations through separate ridge-regression decoders trained to predict mean image-level emotion judgments for neurotypical and autistic participants. The absolute difference between the two population-specific predictions defined the model-predicted diagnostic gap for each image. Layer-wise analyses showed that later layers more reliably generalized the image-level autistic–neurotypical diagnostic gap across model families, whereas earlier layers showed weaker gap predictivity (**Figure S2**). Based on this result, we used feature activations from the penultimate visual layer of each model for the main stimulus-selection analysis (**Figure 2A**). For each model, two test sets were selected: (1) a randomly sampled control set and (2) an ANN-driven set predicted to maximize the NT–ASD difference. These images were subsequently tested on a new cohort of autistic and NT adults. Three outcomes were possible (**Figure 2C**): ANN selection could ($H_0$) make no difference, ($H_1$) worsen separation due to overfitting, or ($H_2$) outperform random selection. The empirical results supported $H_2$.

To test these alternatives, we measured behavioral responses to the selected and randomly sampled images in an independent in-lab cohort of 12 autistic and 13 neurotypical adults (**Figure 3A, for details see Table S1**). Participants performed a two-alternative forced-choice facial emotion judgment task. On each trial, a fixation cross was shown for 1000 ms, followed by a brief face image for 100 ms, a 100-ms blank interval, and a response screen for up to 5000 ms. The response screen contained canonical faces corresponding to the two emotion choices, "happy" and "fearful," and participants indicated which emotion best described the test image. No feedback was provided. The primary behavioral measure was the probability of choosing "happy" for each image. For each stimulus, we averaged responses within the autistic and neurotypical groups and quantified diagnostic power as the absolute difference between group-level response probabilities, |ASD – NT|. This image-level measure allowed us to ask whether images selected prospectively by each ANN produced greater behavioral separation than expected from random image sampling. ANN-selected image sets were compared with same-size random sets matched for identity and emotion-intensity coverage. As an additional control, we repeated the selection procedure after permuting the group labels used to train the population-specific decoders, testing whether diagnostic selection required the true autistic–neurotypical grouping rather than arbitrary partitioning of participants.

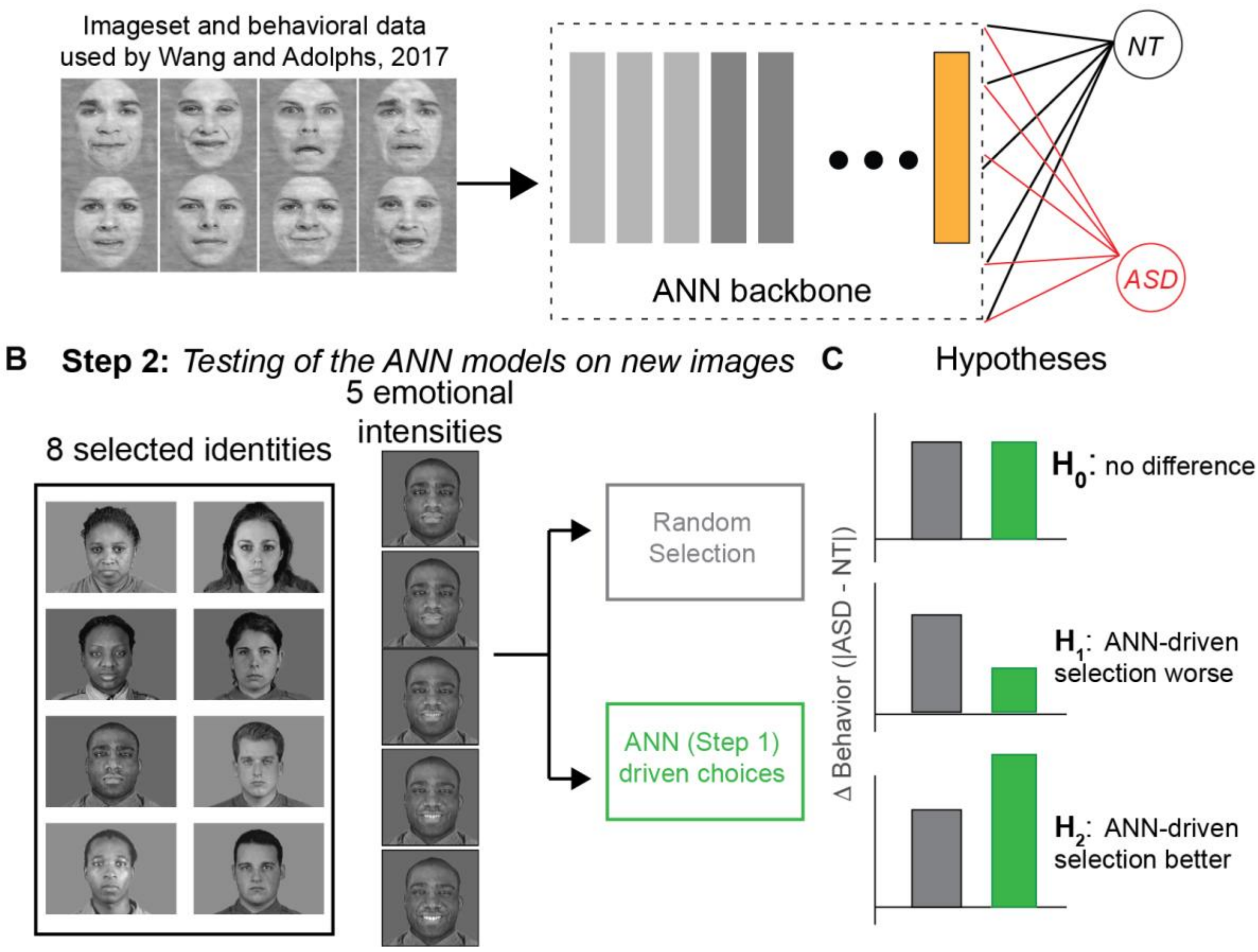

**Figure 2.** Artificial neural networks (ANNs) trained on behavioral data predict diagnostic images in new participants. **A.** *Training ANN models on existing data.* Behavioral responses and facial-image stimuli from *Wang & Adolphs (2017)* were used to train separate neurotypical (NT) and autistic (ASD) behavioral decoders for each ANN architecture. Feature activations from the penultimate layer of each network (AlexNet, VGG-19, ResNet-50, ConvNeXt, CORNet-S, ViT, and CLIP) served as image embeddings. A ridge-regression model was trained to predict the mean probability of judging each face as "happy" for NT and autistic participants, yielding two population-specific prediction streams. The difference between these two predictions (Δ prediction = autistic – NT) quantified the model-predicted behavioral gap for each image. **B.** *Testing the models on a new image set.* To test the predictive power of the trained models, we used the *Montreal Set of Facial Displays of Emotion (MSFDE)*, which contains eight actors expressing five emotional intensities. For each ANN, two subsets of images were selected for testing in new participants: (1) a randomly selected control subset and (2) an ANN-driven subset predicted to yield maximal NT–ASD behavioral difference. The two sets were matched for emotion intensity and identity coverage. **C.** *Hypotheses tested.* Three outcomes were considered: $H_0$ – ANN-driven selection has no effect relative to random sampling; $H_1$ – ANN selection overfits training data and worsens generalization; $H_2$ – ANN-driven selection enhances group separation. Empirical results supported $H_2$.

Random image sets already produced measurable autistic–neurotypical separation, consistent with the idea that the stimulus pool contained diagnostic images but did not distribute them uniformly across images. ANN-selected image sets varied substantially in the extent to which they improved over this random baseline (**Figure 3B**). When we ignored model identity, the average selected-versus-random difference across the seven ANNs was modest and not significant: mean uplift = 0.010 ± 0.012 SEM; paired $t(6) = 0.84$; one-tailed $p = .217$. Thus, simply using an ANN did not guarantee the discovery of diagnostic images. Instead, diagnostic power depended on which model was used to select the images. Model-wise empirical randomization tests showed the clearest improvement for CLIP: CLIP-selected images produced a larger autistic–neurotypical separation than same-size random image sets, with

|ASD–NT| = 0.149 versus a random mean of 0.091; empirical p = 0.006. We next examined the CLIP-selected images more directly to understand how this model improved diagnostic power. Compared with randomly selected images, CLIP-selected images were shifted toward larger signed differences in the probability of choosing "happy" between control and autistic participants, ΔP(happy) = NT − ASD (**Figure 3C**). Thus, CLIP did not merely select an arbitrary subset of faces; it preferentially selected images from the portion of the stimulus distribution that produced stronger group-level behavioral divergence.

This variation across architectures raised a more specific question: what property of a model determines whether its selected images become behaviorally diagnostic? The selection procedure depends on the population-specific predictions generated by each ANN, but those predictions should only be useful if the model captures behaviorally relevant image-level structure in the first place. We therefore asked whether models that better predicted neurotypical image-level judgments were also better at selecting images that separated autistic and neurotypical observers. To quantify this, we measured each ANN's alignment to neurotypical behavior as the correlation between the model's predicted NT judgments and the observed NT responses across images. Models with stronger NT alignment selected image sets that produced larger autistic–neurotypical separation in the independent cohort (**Figure 3D**), Spearman ρ = .82, p = .023. Thus, the success of ANN-guided stimulus discovery depended on whether the model captured behaviorally meaningful image-level variation, rather than on architecture class or model complexity alone.

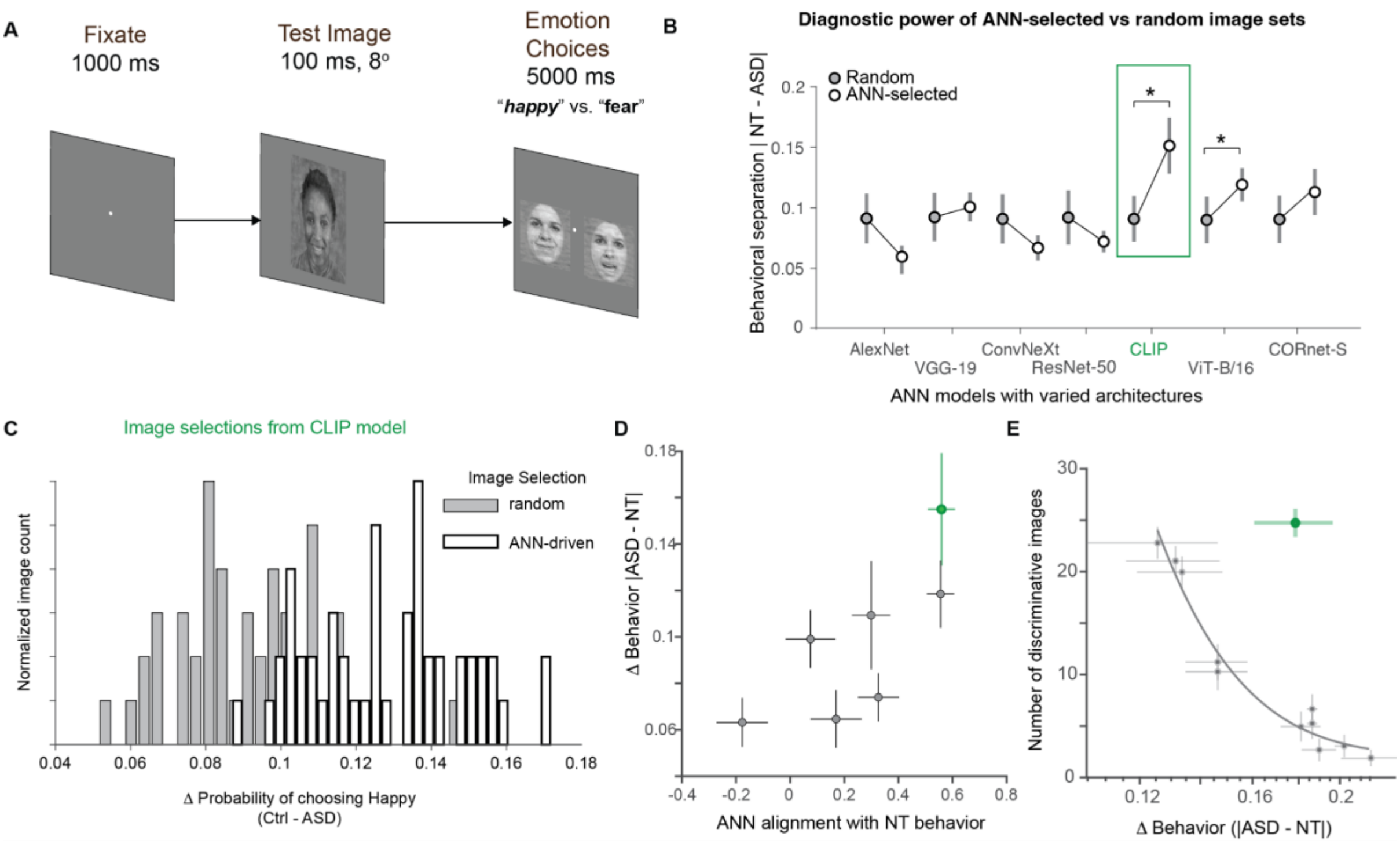

**Figure 3. Behavior-aligned ANNs prospectively identify diagnostic facial-expression stimuli. A.** Behavioral paradigm used to test ANN-selected and randomly selected images in an independent cohort of autistic and neurotypical adults. Each trial began with a 1000-ms fixation, followed by a brief test image presented for 100 ms at an 8° visual angle, and a two-alternative emotion-choice screen. Participants judged whether the face appeared "happy" or "fearful" within a 5000-ms response window. The primary behavioral measure was the probability of choosing "happy" for each image. **B.** Diagnostic power of ANN-selected image sets compared with same-size

random image sets across seven ANN architectures. For each model, images were ranked by the model-predicted absolute difference between autistic and neurotypical emotion judgments, and the top-ranked images were evaluated using observed behavioral responses in the independent cohort. Architectures are ordered by their alignment with neurotypical image-level behavior. Random image sets produced measurable autistic–neurotypical separation, indicating that diagnostic images were present in the stimulus pool, but behaviorally aligned models selected image sets with greater diagnostic power. Across all models, the average selected-versus-random difference was modest when architecture was ignored, with a mean uplift of 0.010 ± 0.012 SEM; paired $t(6)$ = 0.84; one-tailed $p$ = .217. Model-wise empirical randomization tests identified the greatest improvement for CLIP, with selected |ASD–NT| = 0.149 versus random mean = 0.091; empirical $p$ = .006. Asterisks mark model-wise empirical comparisons with $p < .05$. **C.** Distribution of image-level behavioral differences for CLIP-selected images compared with randomly selected images. The x-axis shows the signed difference in the probability of choosing "happy" between control and autistic participants, $\Delta P(\text{happy})$ = Control − ASD. Gray bars show the distribution expected from random image selection, whereas the outlined bars show images selected by the CLIP-based behavioral model. CLIP-selected images were shifted toward larger behavioral differences, indicating that the model preferentially selected images from the high-diagnostic tail of the stimulus distribution. **D.** ANN-selected diagnostic power was predicted by each model's alignment with neurotypical behavior. The x-axis shows the correlation between each model's predicted NT image-level judgments and observed NT behavior; the y-axis shows the observed autistic–neurotypical behavioral separation produced by that model's selected images. Models that better captured NT behavior selected more diagnostic image sets (Spearman ρ = .82, $p$ = .023). **E.** Image-level autistic–neurotypical behavioral differences were sparse. The number of images exceeding a given behavioral-difference threshold decreased sharply as the required difference increased. An exponential fit captured this decline well ($R^2$ = .994), and the relationship between behavioral threshold and the number of discriminative images was strongly negative (Pearson $r$ = −.924, $p = 4.8 \times 10^{-5}$). The highlighted CLIP-selected image set falls in the high-separation region of this distribution, illustrating how behavior-aligned models can identify rare, high-leverage stimuli.

Finally, we examined the distribution of image-level behavioral differences to connect the prospective selection result back to the stimulus-level structure identified in **Figure 1**. In the original Wang and Adolphs dataset, autistic–neurotypical differences were concentrated in a small subset of images rather than distributed uniformly across the stimulus set. The same logic applied to the new MSFDE cohort. Large autistic–neurotypical differences occurred only for a small subset of images: as the required behavioral-difference threshold increased, the number of discriminative images declined sharply (**Figure 3E**). An exponential fit captured this decline well ($R^2$ = .994), and the behavioral threshold correlated strongly and negatively with the number of discriminative images (Pearson r = −.924, $p = 4.8 \times 10^{-5}$). Thus, the images that most strongly differentiated autistic and neurotypical observers again formed a sparse, high-separation tail. Behaviorally aligned ANNs were useful because they prospectively selected images from this tail, rather than relying on post hoc behavioral screening. Together, Figures 1 and 3 show that diagnostic facial-expression stimuli are rare but predictable: image-level group differences are sparse, and behavior-aligned ANN models can identify the high-leverage images most likely to reveal them in new participants.

## Closed-loop synthesis reduces autistic–neurotypical behavioral divergence

The preceding analyses showed that behavior-aligned ANNs can identify facial expressions that produce large autistic–neurotypical differences. We next asked whether the same model-guided framework could transform diagnostic images in a direction predicted to alter the autistic–neurotypical behavioral gap. We focused behavioral validation on gap-reducing synthesis because it provides a conservative, proof-of-concept, intervention test: the algorithm must begin with already diagnostic images and identify realistic transformations that attenuate, rather than merely discover, the measured group difference. This provided a stronger test of the model: if the population-specific predictors captured behaviorally relevant image structure, then changing images in the direction predicted to reduce the model-derived gap should also reduce the measured behavioral gap in new participants.

To test this, we integrated the trained population-specific behavioral models with GANmut, a generative model that represents facial expressions in a continuous two-dimensional latent space. In this space, the angular coordinate controls emotion category and the radial coordinate controls expression intensity, allowing smooth, constrained changes in facial affect while preserving identity and realistic facial structure. For each diagnostic base image, we generated candidate images by updating the GAN latent code through gradient descent. At each iteration, the synthesized image was passed through the ANN-based autistic and neurotypical behavioral models, and the latent code was updated to minimize the predicted difference between the two population-specific "happy" scores (**Figure 4A**). Thus, the optimizer searched the generative image space for a modified version of each face predicted to reduce autistic–neurotypical divergence.

We then tested the resulting synthesized images (see examples, **Figure 4B**) in an independent behavioral cohort (a total of 120 autistic and 51 neurotypical adults recruited on Prolific, **Figure S3**) using a leave-one-image-out phenotype-matched validation analysis. Because the synthesis objective targeted the image-level structure of autistic and neurotypical response profiles, we used correlation-based matching as the primary validation metric. For each held-out image pair, neurotypical participants were selected according to the correlation between their responses to the remaining base images and the original neurotypical behavioral template. Autistic participants were selected in the same way, using the original autistic behavioral template. The held-out image itself was never used for participant selection. This procedure tested whether synthesis reduced the behavioral gap within the response phenotype that the model was trained to capture, while avoiding circularity.

Under this correlation-based phenotype-matched validation, synthesized images reliably reduced autistic–neurotypical behavioral divergence **(Figure 4C**). Across the 15 image pairs, the mean gap decreased from 0.138 for the original base images to 0.076 for the synthesized images, yielding a mean reduction of 0.062 ± 0.026 SEM. Twelve of fifteen image pairs shifted in the predicted direction, $t(14)=2.38$, one-tailed $p=.016$; sign-flip permutation $p=.017$. Thus, model-guided image synthesis attenuated the behavioral separation between autistic and neurotypical observers when validation participants were matched to the image-level response structure that the optimizer targeted.

These results show that the ANN-guided framework can move beyond retrospective prediction and prospective stimulus selection. Starting from diagnostic facial expressions, the closed-loop procedure generated realistic image transformations that reduced autistic–neurotypical divergence in an independent cohort. The effect was not a generic consequence of replacing one image with another; it emerged when behavioral validation was aligned to the population-specific response profiles that guided synthesis.

We next asked whether the gap-reducing effect was restricted to a narrow subset of autistic participants or generalized across trait-defined variation within the online cohort. To test this, we repeated the phenotype-matched validation while grouping autistic participants by SRS and AQ scores, with the neurotypical candidate pool held fixed. Gap reduction remained positive across SRS-defined bins and across most AQ-defined bins (**Figure S4**), indicating that the synthesis effect was not driven by a single idiosyncratic subgroup. However, the magnitude of gap reduction did not vary monotonically with SRS or AQ, suggesting that the targeted image-level phenotype is not simply a proxy for broad autistic-trait severity. Instead, the effect appears to reflect a more specific pattern of image-level emotion judgments that can be present across trait-defined subgroups.

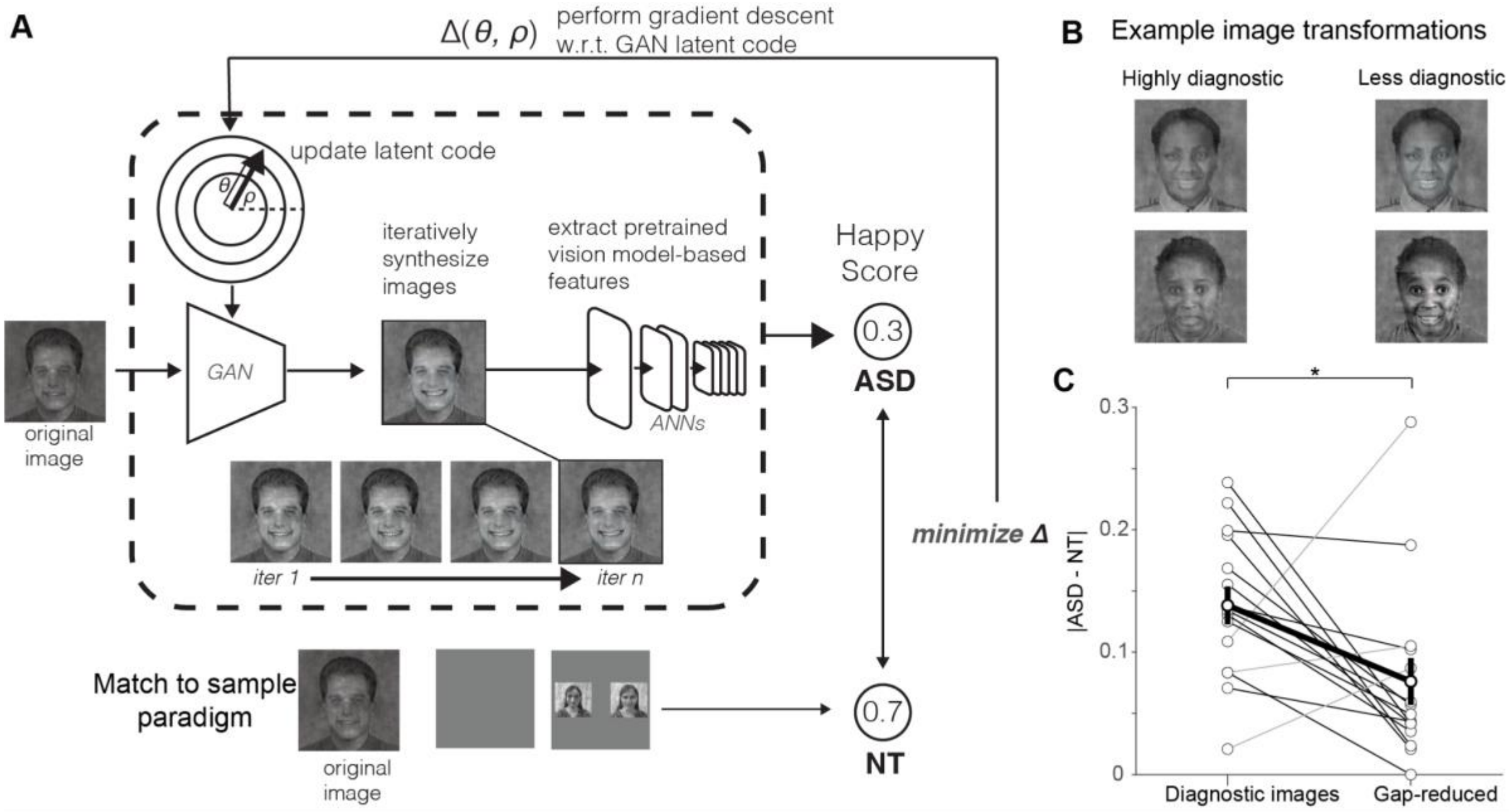

**Figure 4. Closed-loop synthesis reduces autistic–neurotypical behavioral divergence. A.** Closed-loop image-synthesis pipeline for generating gap-reduced facial expressions. Starting from a diagnostic base image, GANmut was used to synthesize candidate facial expressions by updating the latent emotion code. Each candidate image was passed through the trained population-specific ANN behavioral models, which predicted "happy" scores separately for autistic and neurotypical observers. The GAN latent code was then updated via gradient descent to minimize the predicted autistic–neurotypical difference, Δ(θ, ρ), while preserving realistic facial structure and identity. The final synthesized images were tested in the same rapid match-to-sample emotion-discrimination paradigm used for behavioral validation. **B.** Example model-guided image transformations. The left column shows highly diagnostic base images predicted to produce larger autistic–neurotypical behavioral separation. The right column shows corresponding synthesized images predicted to be less diagnostic after closed-loop optimization. **C.** Gap-reducing synthesis reduced autistic–neurotypical behavioral separation in a correlation-based phenotype-matched validation. Each thin line represents an original diagnostic image and its corresponding synthesized image. The y-axis shows the absolute group difference in emotion judgments, |ASD–NT|. Participants were selected using leave-one-image-out phenotype matching: for each held-out image pair, neurotypical and autistic participants were matched to the original group-specific behavioral templates based solely on responses to the remaining base images. Synthesized images reduced the mean gap from 0.138 to 0.076, yielding a mean reduction of 0.062 ± 0.026 SEM. Twelve of fifteen image pairs shifted in the predicted direction, paired one-tailed $t(14)=2.38$, $p=.016$; sign-flip permutation $p=.017$.

## AU analysis describes the facial regions altered by gap-reducing synthesis

We next asked what facial image changes accompanied the model-guided reduction in autistic–neurotypical behavioral divergence. This analysis should be interpreted descriptively, because GANmut constrains image transformations to a two-dimensional expression manifold in which angle controls emotion category and radius controls intensity. Within that constrained space, however, the optimizer could still choose different trajectories for different identities and starting expressions. We therefore summarized each original–synthesized image pair using facial action-unit (AU) regions to ask whether the selected transformations occupied consistent facial regions relative to natural expression-intensity changes. Specifically, we computed a 17-dimensional AU-change vector for each image (**Figure S5**), with each element representing the mean signed pixel change within a single AU-defined facial region (**Figure 5A**). This analysis was not intended to identify a single AU that explained the behavioral effect. Instead, it provided a low-dimensional readout of whether the optimizer altered facial images in structured, reproducible regions of the face.

Across the 15 gap-reducing image pairs, AU-change vectors showed positive similarity for most image pairs (**Figure 5B**). Thus, although each synthesized image preserved the identity and overall structure of its original face, the transformations were not arbitrary. The optimizer repeatedly modulated overlapping combinations of AU-defined facial regions across different images. To quantify this structure, we compared off-diagonal AU-change similarities among synthesized images against a reference distribution derived from natural expression-intensity changes in the original stimulus set. Gap-reducing transformations showed higher AU-region similarity than these natural expression-intensity changes (**Figure 5C**; Welch's $t(93) = 6.22$, $p < .001$), indicating that the closed-loop procedure converged on a coordinated facial-change pattern rather than simply reproducing ordinary increases in happy or fearful expression intensity.

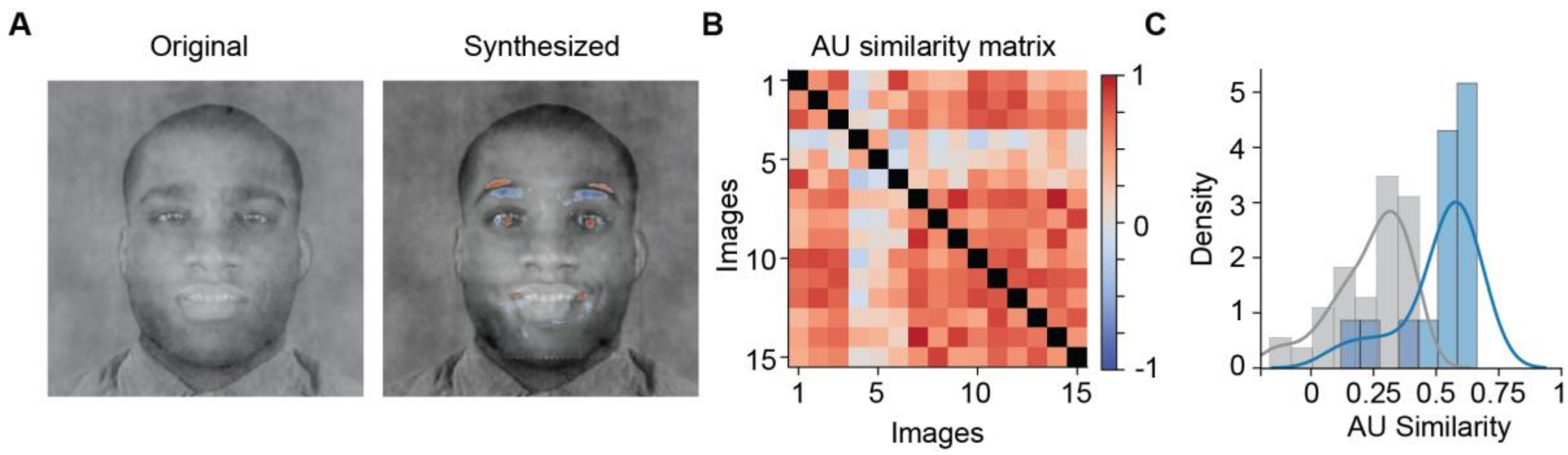

**Figure 5. Gap-reducing synthesis produces structured facial-region transformations.** A. Example of a gap-reducing image transformation. The original diagnostic image is shown on the left, and the synthesized image is shown on the right. Colored overlays indicate pixels that changed relative to the original image after thresholding the absolute pixel-wise difference. Red and blue indicate positive and negative changes in pixel intensity, respectively, highlighting the facial regions most strongly modified by synthesis. B. AU-region similarity matrix for the 15 gap-reducing transformations. For each original–synthesized image pair, pixel-wise changes were summarized within 17 facial action-unit regions to produce an AU-change vector. Matrix entries show pairwise correlations between AU-change vectors across images; the black diagonal indicates self-correlations. Positive off-diagonal similarity indicates that closed-loop synthesis produced partially shared patterns of AU-region modulation across different images. C. Gap-reducing transformations showed structured AU-region changes that differed from ordinary expression-intensity changes. Blue bars show the distribution of AU-change similarities for synthesized images. Gray bars show a reference distribution computed from natural expression-intensity changes in the original stimulus set, using adjacent morph levels from the same identities and emotion categories. Gap-reducing transformations showed higher AU-region similarity than the natural intensity-change reference (Welch's $t(93)=6.22$, $p<.001$), indicating that the optimizer converged on coordinated facial-region changes rather than simply reproducing standard increases in happy or fearful expression intensity.

These AU analyses do not prove that the behavioral effect is caused by a single hand-defined facial feature, nor do they establish that the optimizer discovered a fully unconstrained facial-change pattern. Rather, they provide a descriptive summary of the transformations available within the GANmut manifold. Within that space, gap-reducing synthesis tended to alter overlapping combinations of AU-defined regions while preserving identity and realistic facial structure. We therefore treat the AU results as an interpretability aid, not as an independent mechanistic explanation of autistic–neurotypical behavioral divergence.

# Discussion

Facial emotion perception has long been studied in autism, but group-level effects have often been difficult to convert into sensitive, mechanistically informative assays. Here, we show that this difficulty may arise because autistic–neurotypical differences are sparse rather than uniformly expressed across stimuli. By analyzing individual images, we found that behavioral differences were concentrated in a small subset of high-leverage facial expressions, both in prior data and in our independent test cohort (**Figure 1B, 3E**). This reframes the measurement problem. Instead of asking only whether autistic and neurotypical participants differ on average, we ask which images reveal the difference and whether those images can be predicted or transformed. A sparse image-level phenotype is still a phenotype, but it requires stimulus sets sensitive enough to expose it. This helps explain why facial-emotion findings in autism have been heterogeneous: experiments using the same nominal categories, such as happy and fearful faces, may differ greatly in diagnostic power depending on which images are included. Thus, stimulus variability is not merely nuisance variance; it is part of the phenotype. Model-guided stimulus selection makes this dependence explicit and turns it into an experimentally useful tool. In the sections below, we discuss how behavior-aligned ANNs can be used to discover diagnostic stimuli, how closed-loop synthesis provides a stronger test of model validity, what the resulting image transformations reveal about facial-feature structure, and how this framework can support precision behavioral assay design in autism.

## Behavior-aligned ANNs can help discover diagnostic stimuli

We trained population-specific ANN-based behavioral models on an existing dataset and asked whether they could identify diagnostic images in a new stimulus set before collecting new behavioral data. The answer was not that any ANN could do this (**Figure 3B**). Across architectures, diagnostic stimulus discovery depended on behavioral alignment. Models that better captured neurotypical image-level judgments selected image sets that produced larger autistic–neurotypical behavioral separation in a new cohort. This architecture dependence is important. It argues against a trivial interpretation in which high-dimensional models succeed merely because they can rank images in many arbitrary ways. Successful image selection required the model to capture behaviorally meaningful structure in the visual input. CLIP[42] produced the clearest selection effect in this dataset. Notably, the CLIP variant we used, CLIP-ViT-B/16, shares its vision encoder architecture with one of the other models we evaluated, ViT-B/16. The relative advantage of image-language contrastive models like CLIP in predicting human behavior and neural activity has been attributed to several factors: the larger scale of training data relative to vision-only counterparts[44], the richness of the supervisory signal provided by image-language pretraining[45], and the idea that language encodes the kind of information that is behaviorally relevant even for vision systems[44]. However, the broader principle is that a model's value for assay design depends on its alignment with the relevant human behavior, not simply on its scale, benchmark performance, or architectural class. This places the study within the emerging interface between computational neuroscience, computer vision, and mental health research. Image-computable models have been used to explain visual behavior and neural responses in the primate ventral stream[46–48]. Here, we demonstrate the use of such models in a different role: as engines for experimental design for clinical research[30]. The model is not only asked to explain past responses; it is asked to choose the next stimuli to test. This is a stronger and more translationally useful standard than retrospective prediction because the model must generalize from prior data to a new stimulus set and an independent participant cohort.

## Closed-loop synthesis provides a stronger test of model validity

Stimulus selection asks whether a model can find diagnostic images that already exist in a stimulus set. Synthesis asks whether the model captures enough of the relevant image structure to transform a diagnostic image to achieve a desired behavioral outcome. We therefore integrated the population-specific behavioral models with a generative model of facial expressions and optimized images to reduce the predicted autistic–neurotypical difference. This step turns the model into a closed-loop experimental tool. The model proposes an image transformation, and the transformed image is then tested behaviorally. In a correlation-based, leave-one-image-out phenotype-matched validation, synthesized images reduced the autistic–neurotypical behavioral gap relative to their diagnostic base images (**Figure 4B**). This result shows that the model-guided transformations were not merely visually plausible. They altered the behavioral relationship between groups in the predicted direction. This phenotype-matched validation is critical to this interpretation. The synthesis objective was trained to reduce a particular image-level response profile: the relationship between autistic and neurotypical judgments in the original behavioral data. A new validation cohort may contain participants whose responses only partially reproduce that phenotype. Matching participants to the original group-specific response templates therefore tests the synthesis effect within the behavioral subspace that the optimizer was designed to target, while the leave-one-image-out procedure prevents circularity. Thus, the appropriate claim is not that the synthesized images reduce all possible autistic–neurotypical differences in all samples. Rather, when the targeted image-level phenotype is present, the closed-loop model can identify transformations that attenuate it.

## Gap-reducing transformations are structured but not reducible to a single facial feature

The AU analysis (**Figure 5A-C**) clarifies what kind of transformations the optimizer discovered. The synthesized images did not differ from their base images through arbitrary pixel noise or unrestricted image replacement. Pixel changes were concentrated in facial regions, and AU-region change vectors showed structured similarity across image pairs. At the same time, the effect did not reduce to a single action unit or a simple change in expression intensity. This intermediate result is important. On one hand, the model is not a black box producing uninterpretable random changes. On the other hand, the behavioral effect is not captured by a simple hand-coded rule such as increasing the mouth region, decreasing the brow region, or making all faces more or less happy. Gap-reducing transformations showed coordinated changes across multiple facial regions and differed from natural expression-intensity changes in the original stimulus set. The model therefore appears to identify structured combinations of facial-region changes that depend on the starting image. This supports the broader claim that autistic–neurotypical differences in this task are image-conditioned. The relevant perceptual difference may lie not in whether observers can recognize a category such as happiness or fear, but in how they weigh particular combinations of facial cues under brief viewing conditions[49]. Such combinations are difficult to identify manually but can be discovered using behavior-aligned image-computable models.

## Trait-defined and subgroup analyses support generalization without a simple severity gradient

The dimensional analyses provide an additional constraint on interpretation. When the neurotypical candidate pool was fixed and autistic participants were grouped by SRS or AQ, model-guided gap reduction was generally positive across trait-defined subgroups (**Figure S4**). This suggests that the synthesis effect was not restricted to one idiosyncratic autistic subgroup. However, the magnitude of the effect was not simply explained by SRS or AQ score. This is consistent with the broader heterogeneity of autism[50–52]. SRS and AQ capture meaningful variation in autistic traits, and the online and lab cohorts showed the expected group differences on these measures. However, broad trait scores need not map linearly onto a specific image-level perceptual phenotype. The relevant latent dimension may be more specific: a pattern of responses to particular facial images, a strategy for weighting facial regions, or a sensitivity to particular combinations of local and global facial cues. Thus, the trait analyses support generalization across autistic subgroups without reducing the effect to a simple severity gradient. This distinction is important for clinical and translational interpretation. The goal is not to classify individuals as autistic or neurotypical from facial emotion judgments, nor to define a single autistic facial-emotion phenotype. Rather, the goal is to construct behavioral assays that reveal specific computational dimensions of perception and quantify when those dimensions are expressed.

## Implications for precision behavioral assay design in autism

The present framework suggests a different role for artificial intelligence (AI) in mental health research. The aim is not only to automate diagnosis using various machine learning strategies[53]. Instead, brain-mapped AI models[22] can help design better experiments by identifying the conditions under which a behavioral phenotype is most clearly expressed. It comes with the advantage of providing a neural hypotheses test bed as well since most models' components are mapped to anatomical regions of the brain. In heterogeneous conditions such as autism, this may be especially valuable. A model-guided assay can identify high-leverage stimuli, test whether those stimuli generalize to new participants, what aspects of the stimuli are relevant for its diagnosticity[54], and use synthesis to probe whether the relevant behavioral difference can be modulated through targeted image transformations[55,56].

This approach may complement standard clinical and dimensional measures. Broad instruments such as SRS and AQ summarize trait variation across many behaviors. Image-computable assays target a narrower perceptual computation. These levels of measurement should not be expected to map one-to-one. Instead, they may provide complementary information: trait measures describe broad behavioral phenotype, whereas model-guided stimulus assays identify the image conditions under which a specific perceptual computation diverges or converges. More broadly, the study illustrates how stimulus optimization can improve experimental sensitivity. If a behavioral difference is stimulus-dependent, then the stimulus set should not be treated as a passive background variable. It should be optimized, validated, and, when possible, experimentally perturbed. ANNs make this possible because they can generate image-level predictions for stimuli that have not yet been tested. Generative models extend the approach by allowing researchers to move through stimulus space rather than merely sample from it.

## Limitations

This study should be considered a proof-of-concept, and several limitations should be noted.

First, the behavioral models were trained on group-averaged responses. This was sufficient to identify stimuli and transformations that generalized at the group level, but it does not yet establish individual-level prediction. Future work should build hierarchical or participant-specific models that separate shared group structure from individual variability. Such models could test whether different autistic individuals express distinct image-level phenotypes and whether personalized stimulus selection improves measurement reliability.

Second, the synthesis results depend on the constraints of the generative model. GANmut provides a continuous and interpretable latent space for facial affect, but it can only express transformations available within that learned image manifold. The synthesized images should therefore be interpreted as model-accessible transformations, not as an exhaustive map of all possible features that could reduce autistic–neurotypical differences. Modern diffusion and video-generation models may expand this space, but they will also require stronger controls for identity preservation, realism, demographic bias, and interpretability.

Third, the task was intentionally narrow. Participants made rapid binary judgments along a happy–fearful axis. This design provided a controlled setting in which image-level differences could be quantified, predicted, and perturbed. However, real social perception involves dynamic facial motion, gaze, body posture, vocal affect, context, prior knowledge, and interactive goals[57–59]. Future work should test whether the same model-guided principles extend to richer stimulus spaces, including dynamic expressions and multimodal social cues.

Fourth, the present study does not identify the neural locus of the behavioral differences. Prior work suggests that image-computable models aligned with the primate ventral stream can generate hypotheses about atypical facial emotion processing, including possible roles for inferior temporal cortex and its interactions with affective circuits. The present study builds on that computational logic but does not directly measure neural activity. The image transformations identified here provide a concrete stimulus set for future neural experiments. Functional imaging, eye tracking, intracranial recordings, or nonhuman-primate studies could test whether diagnostic and gap-reduced images differentially engage face-selective visual cortex, amygdala-related affective circuits, or downstream decision systems.

Finally, online recruitment and self-reported diagnostic status introduce additional constraints. We used trait measures and subgroup analyses to assess whether the effects generalized across SRS- and AQ-defined variation, but larger and more deeply phenotyped cohorts will be needed to evaluate clinical generalizability. Future studies should include independent replication samples, broader demographic representation, richer clinical characterization, and explicit tests of whether model-selected stimuli improve reliability relative to conventional stimulus sets.

## Future directions

The approach introduced in this study opens several directions. First, model-guided stimulus selection can be extended to individualized phenotyping. Rather than selecting one diagnostic image set for a population, future experiments could adaptively select stimuli that maximize measurement sensitivity for a specific participant or subgroup. Such closed-loop behavioral testing may be particularly useful in heterogeneous neurodevelopmental conditions. Second, the same approach can be extended beyond static facial expressions. Dynamic faces, audiovisual emotion cues, gaze interactions, and naturalistic social scenes may contain richer diagnostic structure than static images. Image- and video-computable models could be used to identify the temporal and contextual conditions under which autistic and neurotypical perception diverge. Third, combining model-guided stimuli with neural measurements would allow stronger mechanistic tests. Diagnostic and gap-reduced images provide paired stimuli that differ in behavioral effect while preserving many aspects of identity and image structure. These pairs could be used in fMRI, electrophysiology, eye tracking, or causal perturbation experiments to identify neural circuits that track behavioral divergence and convergence. Finally, this framework may generalize beyond autism. Many psychiatric and neurological conditions involve subtle changes in perception, attention, affect, or decision-making that are difficult to capture with standard tasks. Model-guided stimulus discovery could help identify the specific stimulus conditions that reveal these differences, improving assay sensitivity without relying on broad diagnostic labels alone.

## Conclusion

This study shows that autistic–neurotypical differences in facial emotion perception are sparse, image-dependent, and predictable. Behavior-aligned ANNs can prospectively identify facial expressions that produce large group differences, and closed-loop generative synthesis can transform diagnostic images to reduce those differences in a phenotype-matched validation cohort. These findings shift the focus from stimulus-averaged group effects to model-guided discovery of the image conditions under which perception diverges or converges. More broadly, they suggest that AI-guided experimental design can become a powerful tool for precision behavioral phenotyping in autism and related areas of clinical neuroscience.

# Methods

## Participants

Data from human participants were collected with approval from the Caltech Institutional Review Board, and all participants provided informed consent before participating.

### Lab cohort (recruited for ANN-screened behavioral gap increase experiment)

All autistic participants were recruited from the greater Los Angeles area through flyers distributed at various college campuses, outpatient treatment programs, and community support groups, with the following criteria: (a) had a prior diagnosis of autism made by a qualified clinician (M.D. or Ph.D.), (b) autism diagnosis was confirmed through a DSM-5 interview with the participant (and parent, if available) conducted by a licensed clinical psychologist Dr. Lynn K. Paul at Caltech, and (c) scores from ADOS-2 administered at Caltech were either at or above the diagnostic cutoff using the revised scoring algorithm[60]. Autistic individuals were not excluded due to comorbid diagnoses of depression, anxiety, or ADHD, but were excluded if the total score on Beck Depression Inventory-2 fell in the moderate-severe range (25 or above) due to potential confounding effects[61]. Neurotypical control participants were recruited from Caltech's Chen Participant Center participant pool and met the following criteria: age-matched with the autistic cohort, no family history of autism, and no neurological or psychiatric diagnoses.

The lab cohort for testing whether ANN-screening can provide more diagnostic images consisted of 12 autistic participants (2 females, mean age = 32.1, age range = 23-42 years) and 13 NT participants (2 females, mean age = 32.2, age range = 24-39 years). All participants had normal-range IQ and corrected-to-normal visual acuity.

### Online cohort (recruited for closed-loop behavioral gap reduction experiment)

Participants recruited from Prolific[62] met the following eligibility criteria: fluency in English, normal or corrected-to-normal vision, >95% approval rating on Prolific, and a high school diploma or higher level of education completed.

To recruit both autistic and non-autistic participants, we utilized Prolific's prescreening question: "Have you received a formal clinical diagnosis of autism spectrum disorder, made by a psychiatrist, psychologist, or other qualified medical specialist? This includes Asperger's syndrome, Autism Disorder, High Functioning Autism, or Pervasive Developmental Disorder." Participants who responded "Yes - as a child", "Yes - as an adult", or "No" were eligible. Those who provided ambiguous responses (e.g., "I am in the process of receiving a diagnosis", "No, but I identify as being on the autism spectrum", "Don't know/rather not to say") were excluded. We further validated the self-reported diagnoses to reduce the possibility of deception[63] by including the same question regarding clinical diagnoses of autism and other psychiatric disorders after the task. Participants were assured that their responses would remain confidential and would not affect their compensation. In addition to self-reported diagnostic information, we assessed (**Figure S3**) their self-reported autistic traits using the Social Responsiveness Scale-2 (SRS[64]) and the Autism Spectrum Quotient (AQ[65]).

We recruited 120 autistic (52 females, 4 non-binary; mean age = 33.0 ± 10.1, age range = 18-60 years) and 51 neurotypical participants (30 females; mean age = 31.1 ± 10.2, age range = 18-68 years) via Prolific.

# Behavioral task: facial emotion judgment

Participants performed a two-alternative forced-choice (2AFC) facial emotion judgment task designed to measure image-level differences in emotion perception (Figure 3A). On each trial, participants viewed a single face stimulus (test image) and indicated whether the expression appeared *happy* or *fearful*.

Each trial began with a central fixation cross presented for 1000 ms, followed by a test face image presented for 100 ms. In the lab-based experiment, central fixation was ensured with eye-tracking using Tobii Pro Spectrum (600 Hz) before presentation of each test face image, by requiring estimated gaze to be within 10% of screen width and height from the fixation cross. This was followed by a response screen, displaying the two canonical emotion options (happy or fearful), each clearly representing its category. Participants were instructed to select the option that best matched the perceived expression of the face. Responses were collected via keypress, with a maximum response window of 5000 ms (or 3500 ms for the online study). Trials with no response within this window were excluded from further analysis. All trials were presented in randomized order, and no feedback was provided at any point during the task to minimize learning or strategy adaptation.

Stimuli consisted of grayscale images of human faces spanning the continuum from happy to fearful expressions. Across experiments, images varied in identity and emotional intensity. In addition to standard stimulus sets, some experiments included images selected or generated using model-based procedures (described below) to systematically manipulate the expected difference in responses between participant groups.

The primary behavioral measure was the probability of reporting "happy" for each image. For group-level analyses, responses were averaged across participants within each group to obtain an image-level estimate of perceived emotion. Differences between groups were quantified per image as the difference in mean response probability.

### Using ANN-based vision models to screen for diagnostic stimuli

We first constructed a model suite of high-performing deep neural network-based vision models that were representative of performant modern architectures and training objectives. These models included convolutional neural networks (CNNs), specifically AlexNet[36], ResNet-50[38], VGG-19[37], ConvNeXt[39], a vision transformer (ViT[41]), an image-language contrastive model (CLIP[42]), and a CNN designed to be functionally isomorphic to the ventral visual stream (CORNet-S[40]).

Using these models, we developed image-computable pipelines that mapped stimuli (face images) to predicted emotion ratings, ranging from 0 (max fearful) to 1 (max happy), for both autistic and NT populations.

We developed this pipeline in two steps. First, using each model as a visual encoder, we obtained visual embeddings of the stimuli from Wang and Adolphs (2017)[10] by extracting model activations from the penultimate layer in CNNs and ViT and from the `visual embedding' layer in CLIP. Next, we trained cross-validated, regularized ridge regression models to predict the mean emotion rating for a given image (averaged across subjects) from its visual embedding. We trained separate models for autistic and NT subjects.

### Image synthesis pipeline to modulate Autistic–NT predictive differences

Our general approach for synthesizing images that minimized or maximized predicted emotion-rating differences between autistic and neurotypical populations was to integrate our population-specific regression-based prediction models with a pretrained generative adversarial network (GAN). We framed the image synthesis task as an optimization problem, in which the objective was to identify latent codes in the GAN's parameter space that produced images that reduced (or amplified) the predicted rating gap between the two models. We specifically leveraged the pretrained **GANmut[34]** model, which generates images of faces with varied emotions using a two-dimensional latent code.
In this latent code, which is expressed as a polar coordinate space (see Figure 4A), the angle in the space encodes the type of emotion, while the radius encodes its intensity.

### Using GANmut to generate images of facial expressions conditioned on latent codes.

Here, we briefly describe the details of the GAN model (GANmut) used as a part of our image-synthesis pipeline. Standard GAN models used for the conditional generation of faces with varying emotions[66,67] use one-hot codes for different emotion categories. This problem makes it difficult to interpolate between different emotions when generating new images. GANmut remedies this by learning a lightweight, continuous polar latent code based on two parameters: $\theta$ (angle) and $\rho$ (radius). Every possible facial expression can be thought of as a point in this latent space. The generator, *G*, takes as input a latent code and random noise and attempts to generate an image that matches the desired target emotion and intensity based on the regression model's predictions. A standard Discriminator, *D,* attempts to classify the image as real or fake (and a corresponding adversarial loss is computed based on the Discriminator's success). Additional losses, aimed at measuring intensity and emotion type and reconstructing the original latent code used to generate the image, help guide the model during training.

### Image synthesis procedure

#### Reducing the difference in predicted ratings between autistic and neurotypical models for images.

To generate images that minimized the difference between predicted autistic and neurotypical emotion ratings, we used stochastic gradient descent (SGD) to iteratively update the GANmut latent code of a given seed image (see **Figure 4A**). At each iteration, the loss was defined as (neurotypical score − autistic score)$^2$, the squared difference between the neurotypical model's predicted rating of the *original* image and the autistic model's predicted rating of the *synthesized* image. The latent code was updated to minimize this loss until one of two halting criteria was met: the loss fell below a threshold ($\epsilon = .0001$), or 25 iterations were reached. We recorded the final synthesized image, its predicted autistic and neurotypical emotion ratings, the final loss, and the number of iterations required to reach it.

## Action unit based differences between images

To quantify how closed-loop synthesis altered facial images, we summarized pixel-level changes across facial action unit (AU) regions. We used AUCanvas[68], developed by Awakening AI, to identify facial action units (AUs) in the stimulus images. AUCanvas predicts the spatial location and activation probability of 17 facial AUs. For each identity, AUCanvas was applied to the corresponding neutral reference image to define AU-specific spatial regions. These AU regions were then used as fixed masks for comparing the original and synthesized versions of the same identity.

For each original–synthesized image pair, images were first matched in size and converted to the same intensity format. We computed a pixel-wise difference image by subtracting the original image from the synthesized image. For each AU region, we then calculated the mean signed pixel-intensity difference within that region. This yielded a 17-dimensional AU-change vector for each image pair, with each element representing the average change in an AU-defined facial region.

To assess whether synthesis produced similar or heterogeneous facial modifications across images, we compared AU-change vectors across all image pairs. Specifically, we computed pairwise Pearson correlations between the 17-dimensional AU-change vectors for every pair of synthesized images. This produced an AU-change similarity matrix, where higher values indicated that two synthesized images involved similar patterns of AU-region modulation and lower values indicated more image-specific transformations. Self-correlations were excluded from summary statistics.

As a comparison condition, we performed the same AU-based analysis on naturally varying facial expressions from the original stimulus set. For each selected identity, we used images in which the intensity of fearful or joyful expression varied parametrically across morph levels from 20% to 100%. AU-change vectors were computed between matched images of the same identity and emotion category across adjacent intensity levels. These vectors provided a reference distribution of AU-region changes associated with natural increases in facial expression intensity. Pairwise correlations between these reference AU-change vectors were then computed in the same manner as for the synthesized images.

Finally, AU-change similarity values from synthesized images were compared with the reference similarity values obtained from natural expression-intensity changes. This analysis provided a low-dimensional characterization of the facial regions modified by closed-loop synthesis and allowed us to evaluate whether synthesized image transformations followed the same AU-region structure as standard changes in expression intensity.

For comparison, we repeated the analyses using images of the same 8 selected identities in which either a fearful or a joyful facial expression gradually increased (20%, 40%, …, 100%). We again calculated the mean pixel-intensity difference within each AU region between two image pairs in which only the facial expression was slightly increased, and calculated correlations across AUs for all image pairs.

# References

1. Lord, C. *et al.* Autism spectrum disorder. *Nat. Rev. Dis. Primer* **6**, 5 (2020).

2. Behrmann, M., Thomas, C. & Humphreys, K. Seeing it differently: visual processing in autism. *Trends Cogn. Sci.* **10**, 258–264 (2006).

3. Robertson, C. E. & Baron-Cohen, S. Sensory perception in autism. *Nat. Rev. Neurosci.* **18**, 671–684 (2017).

4. Hadad, B.-S. & Yashar, A. Sensory Perception in Autism: What Can We Learn? *Annu. Rev. Vis. Sci.* **8**, 239–264 (2022).

5. Zilbovicius, M. *et al.* Autism, the superior temporal sulcus and social perception. *Trends Neurosci.* **29**, 359–366 (2006).

6. Morrison, K. E. *et al.* Psychometric Evaluation of Social Cognitive Measures for Adults with Autism. *Autism Res.* **12**, 766–778 (2019).

7. Kennedy, D. P. & Adolphs, R. Perception of emotions from facial expressions in high-functioning adults with autism. *Neuropsychologia* **50**, 3313–3319 (2012).

8. Loth, E. *et al.* Facial expression recognition as a candidate marker for autism spectrum disorder: how frequent and severe are deficits? *Mol. Autism* **9**, 7 (2018).

9. Sato, W. *et al.* Impaired detection of happy facial expressions in autism. *Sci. Rep.* **7**, 13340 (2017).

10. Wang, S. & Adolphs, R. Reduced specificity in emotion judgment in people with autism spectrum disorder. *Neuropsychologia* **99**, 286–295 (2017).

11. Beall, P. M., Moody, E. J., McIntosh, D. N., Hepburn, S. L. & Reed, C. L. Rapid facial reactions to emotional facial expressions in typically developing children and children with autism spectrum disorder. *J. Exp. Child Psychol.* **101**, 206–223 (2008).

12. Yi, L. *et al.* Abnormality in face scanning by children with autism spectrum disorder is limited to the eye region: Evidence from multi-method analyses of eye tracking data. *J. Vis.* **13**, 5–5 (2013).

13. Neumann, D., Spezio, M. L., Piven, J. & Adolphs, R. Looking you in the mouth: abnormal gaze in autism resulting from impaired top-down modulation of visual attention. *Soc. Cogn. Affect. Neurosci.* **1**, 194–202 (2006).

14. Dalton, K. M. *et al.* Gaze fixation and the neural circuitry of face processing in autism. *Nat. Neurosci.* **8**, 519–526 (2005).

15. Pelphrey, K. A., Morris, J. P., McCarthy, G. & LaBar, K. S. Perception of dynamic changes in facial affect and identity in autism. *Soc. Cogn. Affect. Neurosci.* **2**, 140–149 (2007).

16. Kleinhans, N. M. *et al.* fMRI evidence of neural abnormalities in the subcortical face processing system in ASD. *NeuroImage* **54**, 697–704 (2011).

17. Rutishauser, U. *et al.* Single-Neuron Correlates of Atypical Face Processing in Autism. *Neuron* **80**, 887–899 (2013).

18. Griffin, J. W. *et al.* Decoding the temporal dynamics of face-specific neural representations in autism. *Nat. Ment. Health* **4**, 1109–1121 (2026).

19. Uljarevic, M. & Hamilton, A. Recognition of Emotions in Autism: A Formal Meta-Analysis. *J. Autism Dev. Disord.* **43**, 1517–1526 (2013).

20. López Pérez, D. *et al.* Visual Search Performance Does Not Relate to Autistic Traits in the General Population. *J. Autism Dev. Disord.* **49**, 2624–2631 (2019).

21. Evers, K., Steyaert, J., Noens, I. & Wagemans, J. Reduced Recognition of Dynamic Facial Emotional Expressions and Emotion-Specific Response Bias in Children with an Autism Spectrum Disorder. *J. Autism Dev. Disord.* **45**, 1774–1784 (2015).

22. Kar, K. & DiCarlo, J. J. The Quest for an Integrated Set of Neural Mechanisms Underlying Object Recognition in Primates. *Annu. Rev. Vis. Sci.* **10**, 91–121 (2024).

23. Schrimpf, M. *et al. Brain-Score: Which Artificial Neural Network for Object Recognition Is Most Brain-Like?* http://biorxiv.org/lookup/doi/10.1101/407007 (2018) doi:10.1101/407007.

24. Rajalingham, R. *et al.* Large-Scale, High-Resolution Comparison of the Core Visual Object Recognition Behavior of Humans, Monkeys, and State-of-the-Art Deep Artificial Neural Networks. *J. Neurosci.* **38**, 7255–7269 (2018).

25. Yamins, D. L. K. *et al.* Performance-optimized hierarchical models predict neural responses in higher visual cortex. *Proc. Natl. Acad. Sci.* **111**, 8619–8624 (2014).

26. Bashivan, P., Kar, K. & DiCarlo, J. J. Neural population control via deep image synthesis. *Science* **364**, eaav9436 (2019).

27. Chang, L. & Tsao, D. Y. The Code for Facial Identity in the Primate Brain. *Cell* **169**, 1013-1028.e14 (2017).

28. Snoek, L. *et al.* Testing, explaining, and exploring models of facial expressions of emotions. *Sci. Adv.* **9**, eabq8421 (2023).

29. Wehrheim, M., Alamooti, S. T., Ramezanpour, H. & Kar, K. Facial expression discrimination emerges from neural subspaces shared with detection and identity. Preprint at https://doi.org/10.1101/2025.08.25.672186 (2025).

30. Kar, K. A Computational Probe into the Behavioral and Neural Markers of Atypical Facial Emotion Processing in Autism. *J. Neurosci.* **42**, 5115–5126 (2022).

31. Yu, H. *et al.* Multimodal investigations of emotional face processing and social trait judgment of faces. *Ann. N. Y. Acad. Sci.* **1531**, 29–48 (2024).

32. Ratan Murty, N. A., Bashivan, P., Abate, A., DiCarlo, J. J. & Kanwisher, N. Computational models of category-selective brain regions enable high-throughput tests of selectivity. *Nat. Commun.* **12**, 5540 (2021).

33. Ponce, C. R. *et al.* Evolving Images for Visual Neurons Using a Deep Generative Network Reveals Coding Principles and Neuronal Preferences. *Cell* **177**, 999-1009.e10 (2019).

34. d'Apolito, S., Paudel, D. P., Huang, Z., Romero, A. & Gool, L. V. GANmut: Learning Interpretable Conditional Space for Gamut of Emotions. in *2021 IEEE/CVF Conference on Computer Vision and Pattern Recognition (CVPR)* 568–577 (IEEE, Nashville, TN, USA, 2021). doi:10.1109/CVPR46437.2021.00063.

35. Tuckute, G. *et al.* How to optimize neuroscience data utilization and experiment design for advancing brain models of visual and linguistic cognition? Preprint at https://doi.org/10.48550/ARXIV.2401.03376 (2024).

36. Krizhevsky, A., Sutskever, I. & Hinton, G. E. ImageNet Classification with Deep Convolutional Neural Networks. in *Advances in Neural Information Processing Systems* (eds Pereira, F., Burges, C. J. C., Bottou, L. & Weinberger, K. Q.) vol. 25 (Curran Associates, Inc., 2012).

37. Simonyan, K. & Zisserman, A. Very Deep Convolutional Networks for Large-Scale Image Recognition. *ArXiv14091556 Cs* http://arxiv.org/abs/1409.1556 (2015).

38. He, K., Zhang, X., Ren, S. & Sun, J. Deep Residual Learning for Image Recognition. Preprint at https://doi.org/10.48550/arXiv.1512.03385 (2015).

39. Woo, S. *et al.* ConvNeXt V2: Co-designing and Scaling ConvNets with Masked Autoencoders. Preprint at https://doi.org/10.48550/ARXIV.2301.00808 (2023).

40. Kubilius, J. *et al.* Brain-Like Object Recognition with High-Performing Shallow Recurrent ANNs. in *Advances in Neural Information Processing Systems* (eds Wallach, H. et al.) vol. 32 (Curran Associates, Inc., 2019).

41. Dosovitskiy, A. *et al.* An Image is Worth 16x16 Words: Transformers for Image Recognition at Scale. Preprint at https://doi.org/10.48550/arXiv.2010.11929 (2021).

42. Radford, A. *et al.* Learning Transferable Visual Models From Natural Language Supervision. Preprint at https://doi.org/10.48550/arXiv.2103.00020 (2021).

43. Beaupré, M. G. & Hess, U. Cross-Cultural Emotion Recognition among Canadian Ethnic Groups. *J. Cross-Cult. Psychol.* **36**, 355–370 (2005).

44. Doerig, A. *et al.* High-level visual representations in the human brain are aligned with large language models. *Nat. Mach. Intell.* **7**, 1220–1234 (2025).

45. Muttenthaler, L., Dippel, J., Linhardt, L., Vandermeulen, R. A. & Kornblith, S. Human alignment of neural network representations. Preprint at https://doi.org/10.48550/arXiv.2211.01201 (2025).

46. Kar, K., Kubilius, J., Schmidt, K., Issa, E. B. & DiCarlo, J. J. Evidence that recurrent circuits are critical to the ventral stream's execution of core object recognition behavior. *Nat. Neurosci.* **22**, 974–983 (2019).

47. Muzellec, S. & Kar, K. Reverse predictivity for bidirectional comparison of neural networks and biological brains. *Nat. Mach. Intell.* **8**, 474–488 (2026).

48. Sörensen, L. K. A., DiCarlo, J. J. & Kar, K. Hierarchical optimization predicts plasticity in the macaque inferior temporal cortex following object training. *Nat. Commun.* https://doi.org/10.1038/s41467-026-74816-0 (2026) doi:10.1038/s41467-026-74816-0.

49. Blais, C., Roy, C., Fiset, D., Arguin, M. & Gosselin, F. The eyes are not the window to basic emotions. *Neuropsychologia* **50**, 2830–2838 (2012).

50. Happé, F., Ronald, A. & Plomin, R. Time to give up on a single explanation for autism. *Nat. Neurosci.* **9**, 1218–1220 (2006).

51. Geschwind, D. H. Advances in Autism. *Annu. Rev. Med.* **60**, 367–380 (2009).

52. Lai, M.-C., Lombardo, M. V. & Baron-Cohen, S. Autism. *The Lancet* **383**, 896–910 (2014).

53. Opel, N. & Breakspear, M. Transforming mental health research and care through artificial intelligence. *Science* **391**, 249–258 (2026).

54. Muzellec, S., Alghetaa, Y. K., Kornblith, S. & Kar, K. MAPS: Masked Attribution-based Probing of Strategies- A computational framework to align human and model explanations. Preprint at https://doi.org/10.48550/arXiv.2510.12141 (2025).

55. Goetschalckx, L., Andonian, A., Oliva, A. & Isola, P. GANalyze: Toward Visual Definitions of Cognitive Image Properties. *ArXiv190610112 Cs* http://arxiv.org/abs/1906.10112 (2019).

56. Younesi, M. & Mohsenzadeh, Y. Controlling Memorability of Face Images. Preprint at http://arxiv.org/abs/2202.11896 (2022).

57. Krumhuber, E. G., Skora, L. I., Hill, H. C. H. & Lander, K. The role of facial movements in emotion recognition. *Nat. Rev. Psychol.* **2**, 283–296 (2023).

58. Aviezer, H., Trope, Y. & Todorov, A. Body Cues, Not Facial Expressions, Discriminate Between Intense Positive and Negative Emotions. *Science* **338**, 1225–1229 (2012).

59. Belin, P., Fillion-Bilodeau, S. & Gosselin, F. The Montreal Affective Voices: A validated set of nonverbal affect bursts for research on auditory affective processing. *Behav. Res. Methods* **40**, 531–539 (2008).

60. Hus, V. & Lord, C. The Autism Diagnostic Observation Schedule, Module 4: Revised Algorithm and Standardized Severity Scores. *J. Autism Dev. Disord.* **44**, 1996–2012 (2014).

61. Dalili, M. N., Penton-Voak, I. S., Harmer, C. J. & Munafò, M. R. Meta-analysis of emotion recognition deficits in major depressive disorder. *Psychol. Med.* **45**, 1135–1144 (2015).

62. Palan, S. & Schitter, C. Prolific.ac—A subject pool for online experiments. *J. Behav. Exp. Finance* **17**, 22–27 (2018).

63. Chandler, J. & Shapiro, D. Conducting Clinical Research Using Crowdsourced Convenience Samples. *Annu. Rev. Clin. Psychol.* **12**, 53–81 (2016).

64. Constantino, J. N. Social Responsiveness Scale. in *Encyclopedia of Autism Spectrum Disorders* (ed. Volkmar, F. R.) 2919–2929 (Springer New York, New York, NY, 2013). doi:10.1007/978-1-4419-1698-3_296.

65. Baron-Cohen, S., Wheelwright, S., Skinner, R., Martin, J. & Clubley, E. The Autism-Spectrum Quotient (AQ): Evidence from Asperger Syndrome/High-Functioning Autism, Males and Females, Scientists and Mathematicians. *J. Autism Dev. Disord.* **31**, 5–17 (2001).

66. Choi, Y. *et al.* StarGAN: Unified Generative Adversarial Networks for Multi-domain Image-to-Image Translation. in *2018 IEEE/CVF Conference on Computer Vision and Pattern Recognition* 8789–8797 (IEEE, Salt Lake City, UT, 2018). doi:10.1109/CVPR.2018.00916.

67. Pumarola, A., Agudo, A., Martinez, A. M., Sanfeliu, A. & Moreno-Noguer, F. GANimation: Anatomically-Aware Facial Animation from a Single Image. in *Computer Vision – ECCV 2018* (eds Ferrari, V., Hebert, M., Sminchisescu, C. & Weiss, Y.) vol. 11214 835–851 (Springer International Publishing, Cham, 2018).

68. Luo, C., Song, S., Xie, W., Shen, L. & Gunes, H. Learning Multi-dimensional Edge Feature-based AU Relation Graph for Facial Action Unit Recognition. in *Proceedings of the Thirty-First International Joint Conference on Artificial Intelligence* 1239–1246 (International Joint Conferences on Artificial Intelligence Organization, Vienna, Austria, 2022). doi:10.24963/ijcai.2022/173.

# Supplementary Material

**AI-guided stimuli discovery and generation to optimize facial emotion perception studies in autism**

**Authors**
Kushin Mukherjee, Na Yeon Kim, Maren Wehrheim, Ralph Adolphs, and Kohitij Kar

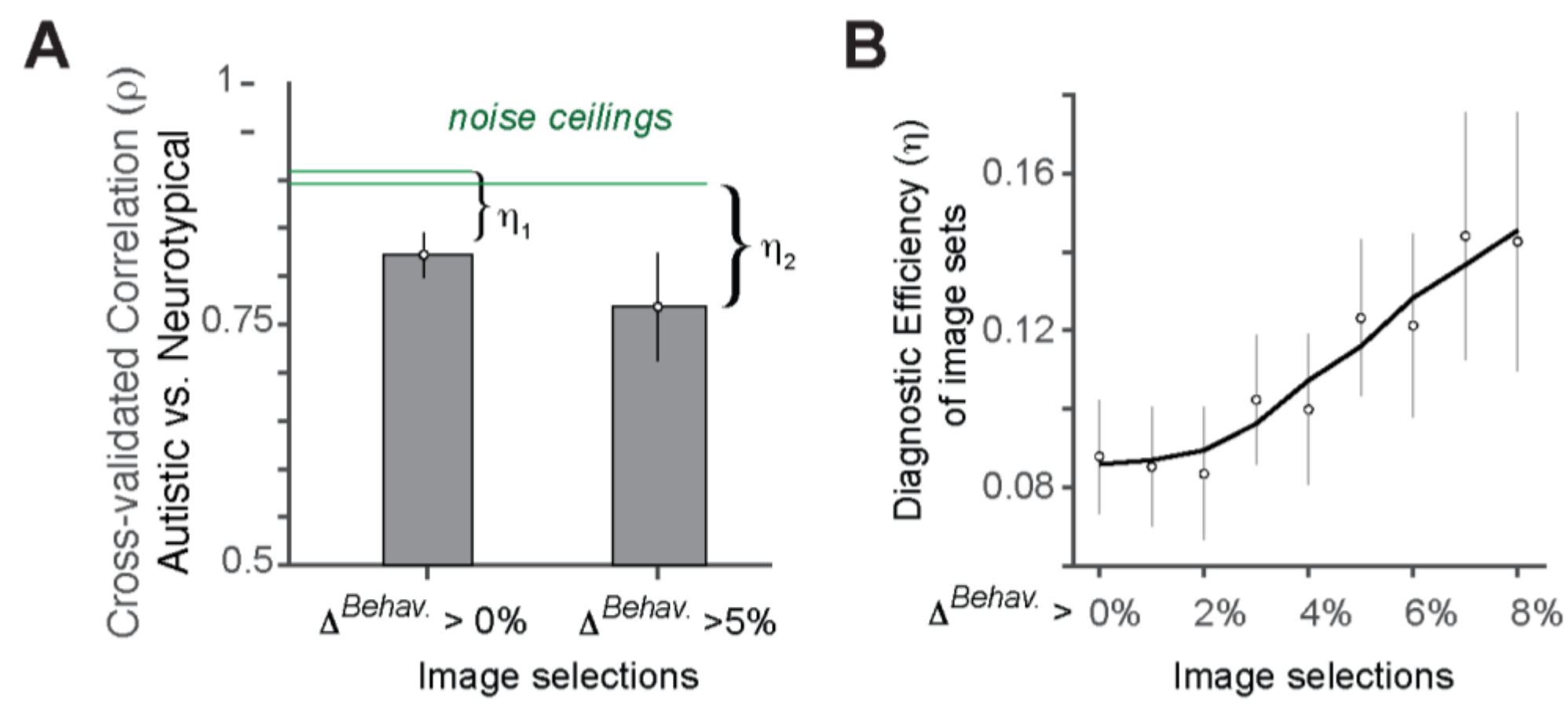

**Figure S1. Image-level behavioral differences can be used to select more diagnostically efficient facial-expression stimuli. A.** Cross-validated correlation between image-by-image behavioral patterns measured in autistic and neurotypical participants for two image-selection criteria. Behavioral patterns were defined as the probability of choosing "happy" for each face image. Image sets were selected using a training subset of subjects according to the image-level behavioral difference between groups, defined as $\Delta\text{Behav} = |P^{\text{Neurotypical}}(\text{happy}) - P^{\text{Autistic}}(\text{happy})|$, and were then evaluated in held-out subjects. Bars show the mean raw cross-validated autistic–neurotypical correlation, ρ, for image sets selected with ΔBehav > 0% and ΔBehav > 5%. Green horizontal lines show the corresponding reliability-limited noise ceilings for each image set. Noise ceilings were estimated from split-half reliability of the image-level behavioral vectors separately within each group, corrected using the Spearman–Brown procedure, and combined across groups to estimate the maximum expected autistic–neurotypical correlation given measurement noise. The difference between the noise ceiling and the observed raw correlation defines the diagnostic efficiency of the image set, η. Larger η indicates that an image set more efficiently separates autistic and neurotypical image-level behavioral patterns beyond what can be explained by measurement noise. **B.** Diagnostic efficiency, η, as a function of increasingly stringent image-selection criteria. The x-axis shows the minimum required image-level behavioral difference, ΔBehav, used to select images in the training subjects, and the y-axis shows diagnostic efficiency measured in held-out subjects. Error bars indicate bootstrap confidence intervals across subject-level cross-validation and image-selection iterations. Diagnostic efficiency increased as images were selected using larger autistic–neurotypical behavioral differences, demonstrating that image-level stimulus selection can produce more powerful behavioral assays for detecting atypical facial emotion judgments.

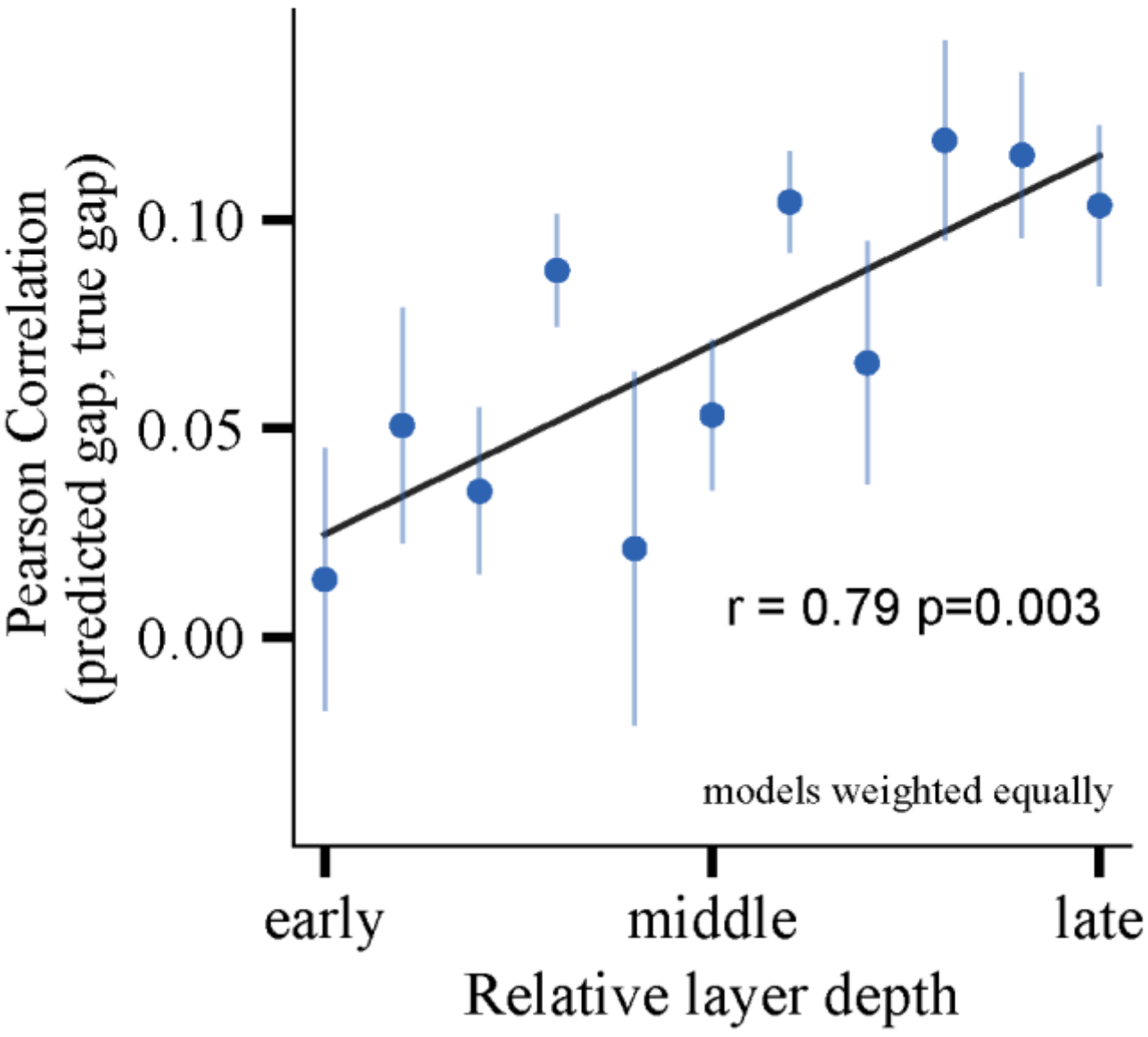

**Figure S2. Layer depth predicts model sensitivity to ASD–NT behavioral gaps.** For each neural network model and sampled layer, we evaluated how well the model-predicted diagnostic gap generalized to the held-out MSFDE image set. The diagnostic gap was defined image-by-image as the absolute difference between the predicted autistic and neurotypical/control responses, and was compared against the corresponding observed behavioral gap. Each blue point shows the median Pearson correlation between predicted and observed image-level gaps at a given relative layer depth, after first aligning layers from different architectures onto a common normalized depth axis from early to late processing stages. Early, middle, and late denote low-, intermediate-, and high-level representational stages, respectively. Models were weighted equally so that architectures with more sampled layers did not dominate the average trend. Error bars indicate median absolute deviation divided by the square root of the number of models, providing a robust estimate of variability across model families. The black line shows the linear fit across layer depth. The positive correlation (Pearson r = 0.79, p=0.003) indicates that later model layers more reliably capture the image-by-image structure of the behavioral gap than earlier layers. Thus, across architectures, higher-level visual representations show stronger generalization to diagnostic differences in emotion judgments.

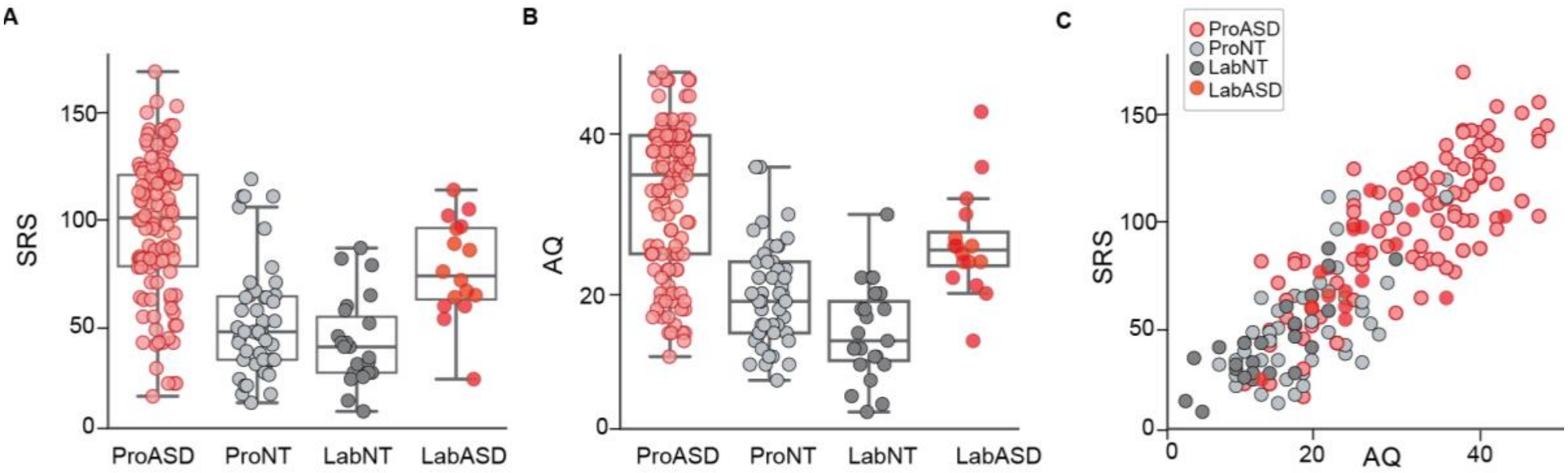

**Figure S3.** Comparison of autistic traits across the lab cohort (LabASD and LabNT) and online cohort recruited via Prolific (ProASD and ProNT). **A.** Based on the total scores of the Social Responsiveness Scale (SRS), both LabASD and ProASD groups showed significantly higher autistic traits than LabNT (adjusted $p$ = 0.003) and ProNT (adjusted $p < 0.001$) groups, respectively. **B.** The total scores of the Autism Spectrum Quotient (AQ) showed the same pattern. C. Both measures of autistic traits were highly correlated ($r$ = 0.85, $p < 0.001$), further supporting the validity of the self-reported measures across all groups.

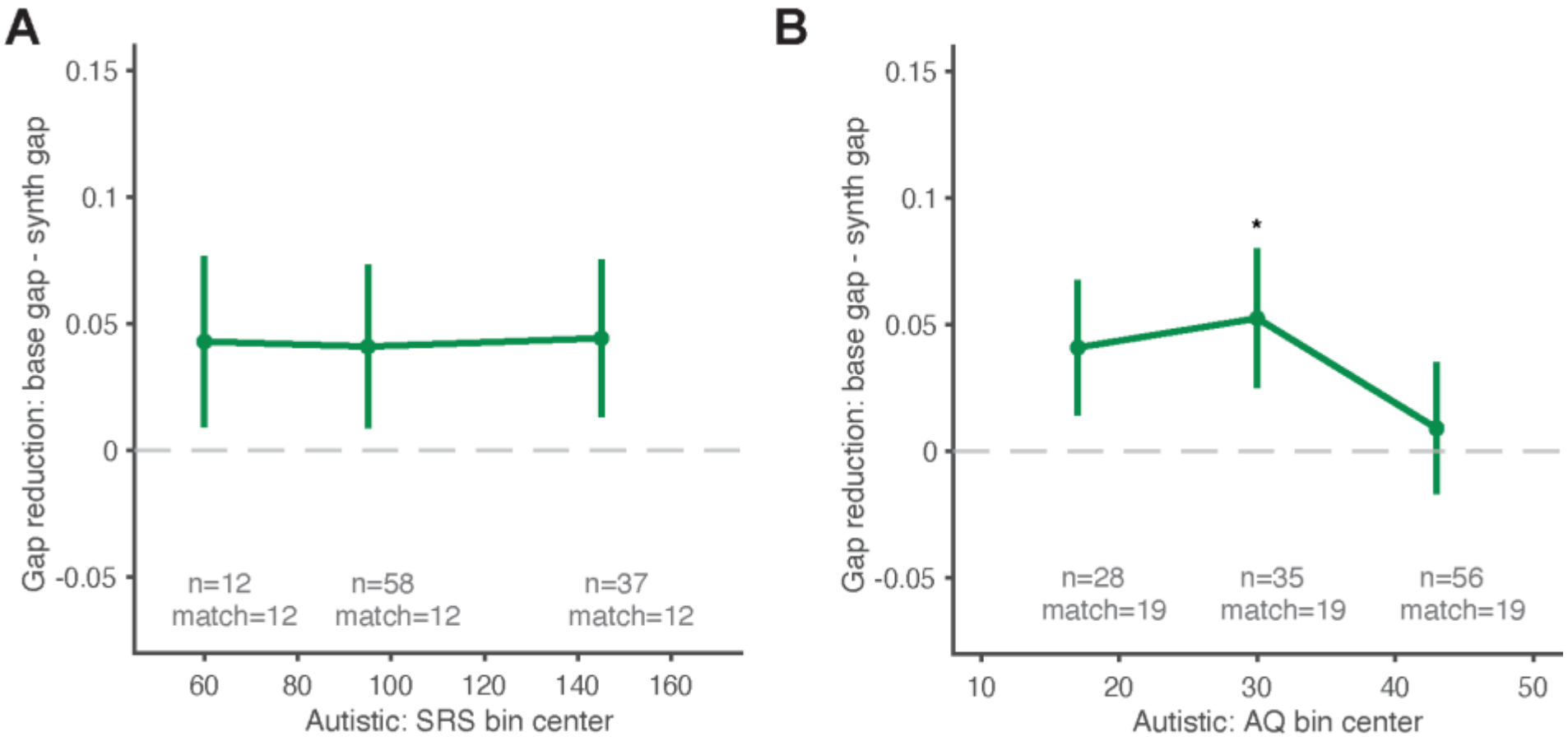

**Figure S4. Gap-reducing synthesis generalizes across autistic trait-defined subgroups. A.** Gap reduction for ASD participants grouped by SRS score. The neurotypical candidate pool was fixed across all comparisons and consisted of Control participants with SRS < 50 and AQ < 20. For each SRS bin, ASD participants were selected from that bin, and both neurotypical and ASD subjects were matched to the original group-specific behavioral templates using leave-one-image-out correlation-based phenotype matching. The same 15 diagnostic base images and their matched gap-reduced synthesized images were used in all bins. Positive values indicate that synthesized images reduced the autistic–neurotypical behavioral gap relative to the original diagnostic images. Error bars indicate SEM across image-level gap-reduction effects. **B.** Same analysis for ASD participants grouped by AQ score. Across AQ-defined subgroups, gap reduction remained positive on average, indicating that the synthesis effect was not restricted to a single ASD trait subgroup. Together, these analyses show that model-guided gap reduction generalized across participants with different SRS and AQ scores, while the magnitude of the effect was not simply explained by autistic-trait severity.

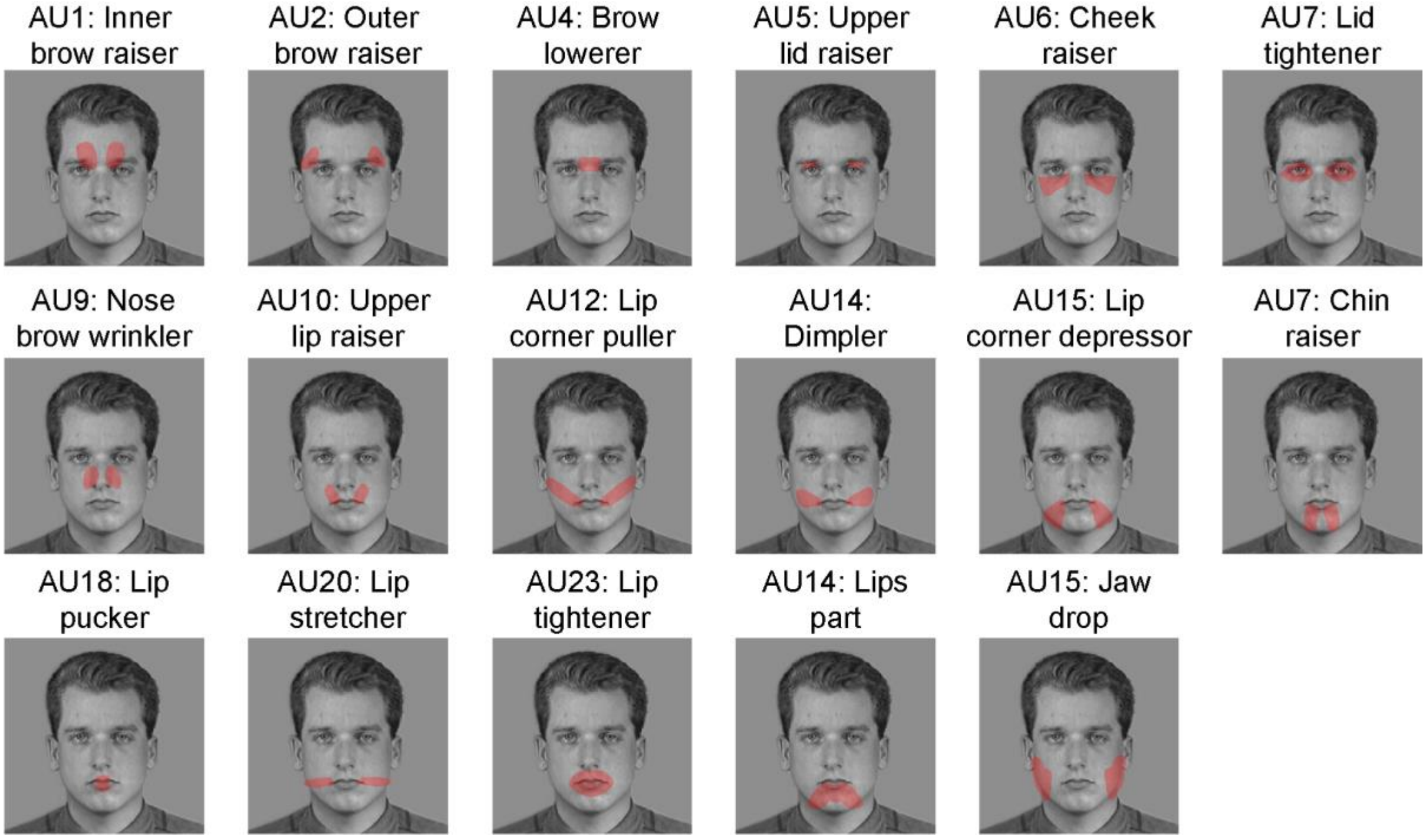

**Figure S5. Facial action-unit regions used to summarize image transformations.** Representative facial action-unit (AU) regions used for the AU-based analysis of original–synthesized image differences. A neutral reference face is shown with AU-specific spatial masks overlaid in red. Each panel illustrates one AU-defined facial region, including brow, eyelid, cheek, nose, mouth, lip, chin, and jaw regions. For each identity, AU regions were estimated from the corresponding neutral reference image and then used as fixed spatial masks for comparing the original and synthesized versions of that identity. In the image-change analysis, pixel-wise differences between the original and synthesized images were averaged within each AU region, producing a 17-dimensional AU-

change vector for each image pair. These vectors were then compared across images to quantify whether closed-loop synthesis produced shared or image-specific patterns of facial-region modulation.

| Grp | Sub | Age | Sex | Hand | Race | Education | ADOS CSS-overall | FSIQ | PIQ | VIQ | SRS | AQ |
|---|---|---|---|---|---|---|---|---|---|---|---|---|
| Autistic | 1 | 42 | M | R | Caucasian | Bachelor's Degree | 8 | 91 | 111 | 50 | 72 | 26 |
| | 2 | 42 | M | L | Caucasian | Bachelor's Degree | 9 | 99 | 101 | 97 | 64 | 36 |
| | 3 | 25 | F | n.a. | Asian | Master's Degree | 4 | n.a. | n.a. | n.a. | 96 | 25 |
| | 4 | 41 | M | R | Asian/Pacific Islander | Master's Degree | 8 | 124 | 115 | 127 | 97 | 26 |
| | 5 | 30 | M | R | Hispanic/Latino | Some College | 8 | 108 | 103 | 110 | 76 | 21 |
| | 6 | 30 | M | A | Asian | Some College | 9 | 125 | 119 | 123 | 65 | 22 |
| | 7 | 36 | M | R | Caucasian | High School | 7 | 93 | 99 | 89 | 105 | 32 |
| | 8 | 25 | M | R | Caucasian | Some College | 7 | 114 | 111 | 113 | 89 | 30 |
| | 9 | 27 | F | R | Caucasian | Master's Degree | 7 | 116 | 109 | 119 | 86 | 26 |
| | 10 | 23 | M | n.a. | n.a. | Some College | n.a. | n.a. | n.a. | n.a. | n.a. | n.a. |
| | 11 | 33 | M | R | Caucasian | Associate's Degree | 10 | 100 | 109 | 90 | 54 | 24 |
| | 12 | 31 | M | R | Caucasian | Some College | 9 | 104 | 117 | 90 | 67 | 24 |
| Neurotypical | 1 | 31 | M | n.a. | n.a. | n.a. | - | n.a. | n.a. | n.a. | 60 | 17 |
| | 2 | 39 | F | R | Caucasian | Bachelor's Degree | - | 116 | 111 | 116 | 11 | 7 |
| | 3 | 37 | F | L | Hispanic/Latino | Bachelor's Degree | - | 100 | 94 | 106 | 52 | 18 |
| | 4 | 26 | M | R | Asian | Master's Degree | - | n.a. | n.a. | n.a. | 46 | 18 |
| | 5 | 30 | M | R | African American / Black | Bachelor's Degree | - | n.a. | n.a. | n.a. | 26 | 13 |
| | 6 | 33 | M | R | Hispanic/Latino | Associate's Degree | - | 93 | 84 | 103 | 41 | 9 |
| | 7 | 27 | M | L | Caucasian | Bachelor's Degree | - | n.a. | n.a. | n.a. | 79 | 22 |
| | 8 | 24 | M | A | Asian | Bachelor's Degree | - | n.a. | n.a. | n.a. | 65 | 20 |
| | 9 | 33 | M | R | Hispanic/Latino | Bachelor's Degree | - | n.a. | n.a. | n.a. | 36 | 6 |
| | 10 | 25 | M | R | Caucasian | Associate's Degree | - | n.a. | n.a. | n.a. | 58 | 22 |
| | 11 | 35 | M | R | Caucasian | Master's Degree | - | n.a. | n.a. | n.a. | 31 | 11 |
| | 12 | 39 | M | R | Caucasian | Bachelor's Degree | - | n.a. | n.a. | n.a. | 87 | 22 |
| | 13 | 39 | M | A | Hispanic/Latino | Some College | - | n.a. | n.a. | n.a. | 41 | 20 |

**Table S1. Characterization of the lab cohort.** Age at experiment. Hand: Dominant handedness (A: ambidextrous, L: left, R: right). Education: Highest education level completed. ADOS CSS-overall: Autism Diagnostic Observation Schedule, calibrated severity scores, which were generated from the Hus and Lord algorithm (ASD group only) (Hus & Lord, 2014). FSIQ, PIQ, VIQ: Full-scale Intelligence Quotient (FSIQ), performance IQ (PIQ), verbal IQ (VIQ) scores from the Wechsler Abbreviated Scale of Intelligence, 2nd edition (Wechsler, 2011). SRS: Social Responsiveness Scale-2 (Constantino, 2012) Adult Form (Self report), total score. AQ: Autism Spectrum Quotient (Baron-Cohen et al., 2001), total score. For ADOS CSS-overall, SRS, and AQ, higher scores indicate stronger autism traits. n.a.: not available.